\documentclass[preprint,12pt,authoryear,]{elsarticle}
\usepackage{amssymb}
\usepackage[normalem]{ulem}

\usepackage{booktabs}
\usepackage{float}
\usepackage{caption}
\usepackage{subfig}

\usepackage{xcolor} 
\definecolor{Reviewer1Color}{rgb}{0.0, 0.5, 0.0} 
\definecolor{Reviewer2Color}{rgb}{0.0, 0.0, 0.8} 
\definecolor{RemovedTextColor}{rgb}{0.8, 0.0, 0.0} 
\definecolor{ChangesColor}{rgb}{0.48, 0.25, 0.52} 

\setcitestyle{authoryear,open={(},close={)}}

\journal{International Review of Economics and Finance.}
\begin{document}

\begin{frontmatter}


\title{FinTextSim: Enhancing Financial Text Analysis with BERTopic}

\affiliation[inst1]{organization={Universidad Politécnica de Madrid, DEGIN doctoral program, Department of Industrial Management. ETSII},
            addressline={c. José Gutiérrez Abascal 2}, 
            city={Madrid},
            postcode={28006}, 
            state={Madrid},
            country={Spain}}

\affiliation[inst2]{organization={Fakultät für Wirtschaft, Hochschule Heilbronn},
           addressline={Bildungscampus, Max-Planck-Straße 39}, 
           city={Heilbronn},
           postcode={74081}, 
           state={State Two},
           country={Germany}}

\affiliation[inst3]{organization={Escuela Técnica Superior de Ingeniería Industrial, Universidad de la Rioja},
           addressline={C. Luis de Ulloa, 4}, 
           city={Logroño},
           postcode={26004}, 
           state={La Rioja},
           country={Spain}}

\affiliation[inst4]{organization={Beta Klinik GmbH},
            addressline={Joseph-Schumpeter-Allee 15}, 
            city={Bonn},
            postcode={53227}, 
            country={Germany}}


\author[inst1,inst4]{Simon Jehnen}

\author[inst2,inst3]{Javier Villalba-D\'iez}

\author[inst1]{Joaqu\'in Ordieres-Mer\'e}

\begin{abstract}
Recent advancements in information availability and computational capabilities have transformed the analysis of annual reports, integrating traditional financial metrics with insights from textual data. 
To extract valuable insights from this wealth of textual data, automated review processes, such as topic modeling, are crucial. 
This study examines the effectiveness of BERTopic, a state-of-the-art topic model relying on contextual embeddings, for analyzing Item 7 and Item 7A of 10-K filings from S\&P 500 companies (2016–2022).
Moreover, we introduce FinTextSim, a finetuned sentence-transformer model optimized for clustering and semantic search in financial contexts. 
Compared to all-MiniLM-L6-v2, the most widely used sentence-transformer, FinTextSim increases intratopic similarity by 81\% and reduces intertopic similarity by 100\%, significantly enhancing organizational clarity.  
We assess BERTopic’s performance using embeddings from both FinTextSim and all-MiniLM-L6-v2.
Our findings reveal that BERTopic only forms clear and distinct economic topic clusters when paired with FinTextSim's embeddings. 
Without FinTextSim, BERTopic struggles with misclassification and overlapping topics.
Thus, FinTextSim is pivotal for advancing financial text analysis.
FinTextSim’s enhanced contextual embeddings, tailored for the financial domain, elevate the quality of future research and financial information.
This improved quality of financial information will enable stakeholders to gain a competitive advantage, streamlining resource allocation and decision-making processes. 
Moreover, the improved insights have the potential to leverage business valuation and stock price prediction models.
\end{abstract}



\begin{keyword}
Topic Modeling \sep 10-K \sep Artificial Intelligence \sep BERTopic 
\sep MD\&A \sep FinTextSim \sep Sentence Transformers
\end{keyword}

\end{frontmatter}

\section*{Acknowledgement}
The second and third authors want to acknowledge the partial support by the Spanish “Agencia Estatal de Investigación” through the grant PID2022-137748OB-C31 funded by MCIN/AEI/10.13039/501100011033 and "ERDF A way of making Europe”.

\section{Introduction}
In recent years, the increasing availability of information \citep{gupta2020,ABUKARI2024103433} and advances in computational capabilities have transformed the way we analyze annual reports, including 10-K filings.
10-K filings are among the most critical annual reports \citep{griffin2003}, providing a standardized snapshot of a company's financial situation through both numerical and textual data \citep{masson2020}. 
The information contained carries predictive power for future profitability \citep{you2009}.
Hence, various stakeholders, such as investors and financial analysts, rely on 10-K reports to make informed decisions \citep{liu2022}.
Traditionally, evaluation focused on retrospective quantitative financial metrics. However, there is a growing recognition of the value embedded in qualitative textual data.
For example, \citet{cohen2020,li2010a} demonstrated that language and tone in financial reports correlate with future company returns. 
Therefore, integrating retrospective financial metrics with textual analysis offers a fuller picture of a company, improving various decision-making processes \citep{hsieh2022}.

Item 7 and Item 7A of 10-K filings contain valuable information regarding companies listed in the Standard and Poor's 500 (S\&P 500). 
Among the 15 items included in 10-K reports, they stand out as particularly crucial.
Item 7 is the Management Discussion \& Analysis (MD\&A) section.
In this section, the management presents the company's perspective on various aspects, including operations, performance, risks, opportunities, and strategies to address future challenges \citep{cohen2020}. 
Item 7A contains qualitative and quantitative disclosures about market risk.
As the submission of a 10-K filing is mandatory for publicly traded companies, there is a wealth of information. We need clearly defined review processes to evaluate and use this information~\citep{DUTTA2019130}. 

Automated review processes offer various advantages over manual approaches. 
First, manual review of textual data is time-consuming and prone to subjectivity bias \citep{li2010a,li2010}. 
Moreover, the vast and growing amount of data \citep{gupta2020} can lead to information overload, particularly for stakeholders with limited attention \citep{lu2022}. Thus, attention must be efficiently allocated \citep{liu2022}. 
Topic modeling, a technique from Natural Language Processing (NLP), addresses these challenges by uncovering latent topics within textual datasets. Hence, they help to organize and summarize large text corpora \citep{blei2003}.

Recently developed neural topic models address the limitations of classical topic modeling approaches, which still dominate applied topic modeling.
Classical topic modeling approaches have been discussed in the literature since the introduction of Latent Semantic Indexing in 1990 \citep{deerwester1990}. 
Until 2015, Bayesian probabilistic models, most notably Latent Dirichlet Allocation (LDA), were considered state-of-the-art. 
Relying on the bag-of-words (BoW) assumption, each document is treated as a collection of words, disregarding their sequential order. However, this approach limits the model's ability to capture the semantic meaning of text.
Neural topic modeling approaches address this issue by employing contextual embeddings \citep{blair2020}, allowing them to capture richer semantic and contextual relationships within the data \citep{booker2024,bhattacharya2024}.

Recent advances in contextual embeddings have transformed NLP, driven by key innovations such as the transformer architecture, encoder-only models, and sentence-transformers.
The transformer architecture, introduced by \citet{vaswani2017}, revolutionized NLP by relying entirely on attention mechanisms, allowing models to capture long-range dependencies and rich contextual information. 
This made transformers the state-of-the-art approach for Natural Language Understanding tasks~\citep{caron2020,HAN2024103508}.
Building on this foundation, Bidirectional Encoder Representations from Transformers (BERT) established a new standard for deep contextualized language modeling \citep{devlin2019}. 
BERT relies on the encoder from the transformers architecture, allowing it to processes text bidirectionally and capture nuanced semantic relationships. 
More recently, \citet{warner2024} introduced improvements to this architecture, increasing efficiency and surpassing BERT in classification and retrieval tasks.
Despite their strengths, encoder-only models are less effective for large-scale semantic similarity comparison and clustering \citep{reimers2019}. 
To address these limitations, sentence-transformers refine encoder-only models using siamese or triplet network architectures, enabling efficient and precise similarity assessments \citep{reimers2019}. 
Sentence-transformers encode text into dense vector representations—contextual embeddings—that quantify semantic similarity by mapping similar texts closer in a shared vector space.
Neural topic models, such as BERTopic, leverage these contextual embeddings to enhance topic modeling by clustering semantically related documents, enabling more accurate topic discovery \citep{grootendorst2022}.

While these advancements have demonstrated significant improvements in general text processing, it remains unclear how these sophisticated methods perform when applied to tasks specifically relevant to the finance and accounting domains \citep{bhattacharya2024}. Additionally, there is little evidence that the customization of financial text leads to benefits in performance \citep{huang2023}.
Traditional algorithms continue to dominate applied topic modeling, hindering the generation of new knowledge \citep{egger2021, blair2020,huang2023}.
Although extensive studies on various topic modeling approaches \citep{albalawi2020, fu2021, egger2022, farzadnia2024} have been conducted, the domain of Management Accounting and Finance, particularly Item 7 and Item 7A of 10-K reports from S\&P 500 companies, remains significantly under-researched. 
This presents a critical opportunity to integrate Machine Learning (ML)-based methods to fully exploit the value hidden in financial textual data \citep{ranta2022}. 
Furthermore, it opens avenues for refining domain-specific contextual embeddings by incorporating domain expertise \citep{murphy2024}. 
Such an approach aims to enhance the quality of contextual embeddings, allowing them to capture terminology and concepts unique to finance with greater precision \citep{dong2024}.

To address this gap, we introduce FinTextSim, a finetuned sentence transformer leveraging the capabilities of contextual embeddings for the financial domain.
We benchmark FinTextSim against all-miniLM-L6-v2 (AM), the most widely used general-purpose sentence transformer. 
To isolate the effect of the selected sentence-transformer, we generate contextual embeddings for our dataset using both FinTextSim and AM.
We apply these embeddings to BERTopic, a state-of-the-art neural topic modeling approach, while keeping all other model parameters identical.
This comparison allows us to determine whether domain-specific fine-tuning enhances topic modeling and financial text interpretation.

We will explore the following research questions based on Item 7 and Item 7A from S\&P500 companies between 2016 and 2022:
\begin{enumerate}
    \item[RQ1] How can we leverage the capabilities of contextual embeddings for the financial domain? 
    \item[RQ2] Which embedding model — FinTextSim or AM — produces more qualitative and coherent topics when used as input for BERTopic?
    \item[RQ3] Which embedding model — FinTextSim or AM — better supports BERTopic in organizing and summarizing large-scale financial text corpora?
\end{enumerate}

By addressing the research questions, our work makes the following significant contributions:
\begin{enumerate}
    \item We identify the most effective approach for extracting meaningful topics from Item 7 and Item 7A of S\&P500 companies. This comparison provides valuable insights for researchers and practitioners selecting embedding models for NLP tasks.
    \item We introduce FinTextSim, a finetuned sentence-transformer, improving the analysis of financial text for various downstream tasks. Hence, FinTextSim will boost future research quality.
    \item By enhancing BERTopic for financial text with FinTextSim, we generate higher quality financial information regarding companies, sectors as well as whole markets and economies. Thus, we enable managers, financial analysts, investors, regulators and other stakeholders to gain a competitive advantage which aids in allocating resources more efficiently and making rational operational and strategic decisions. Moreover, the improved insights have the potential to leverage business valuation and stock price prediction models.
\end{enumerate}

The rest of the paper hereinafter is organized as follows. Section 2 reviews the state-of-the-art literature and methodologies. Section 3 describes our study's materials and methods, including the training procedure of FinTextSim. Section 4 presents and discusses the main findings. Finally, Section 5 provides the conclusion. This structure ensures a clear and logical progression, enabling a thorough understanding of our study's contributions.

\section{State of the Art}
The following subsections provide an overview of the evolution of contemporary topic modeling techniques and a detailed examination of BERTopic, a state-of-the-art topic modeling approach.

\subsection{Evolution of Contemporary Topic Modeling Approaches}
Recent advancements in topic modeling have seen the integration of contextual embeddings, offering both significant benefits and challenges. 
Modern methodologies address the limitations of classical models by utilizing advanced text embedding techniques, moving beyond simple BoW representations. This enables them to better capture semantic relationships within text \citep{blair2020}.
While classical models require extensive customization and become increasingly complex with larger datasets, modern approaches offer enhanced flexibility and scalability \citep{zhao2021,wu2024}. 
Moreover, neural topic models simplify the inference problem, enabling parallelization \citep{abdelrazek2023,wu2024}.
Among other innovative methods \citep{wu2024}, contextual vector representations are combined with centroid-based clustering techniques \citep{sia2020, angelov2020}.
They assume that the centroid of a cluster represents the topic. Words closest to the centroid are considered as the most representative ones for the topic. 
However, this assumption is fragile as clusters may not always conform to a spherical distribution around the centroid. As a result, misrepresentation of topics may occur \citep{grootendorst2022}.
A promising approach for topic modeling based on contextual embeddings, addressing centroid-based issues, is BERTopic.

\subsection{BERTopic}
BERTopic structures topic modeling into five sequential steps. 
First, document embeddings are generated using a pre-trained sentence transformer. 
A method that offers enduring benefits by leveraging advancements in language models \citep{grootendorst2022,gu2024a}.
Second, the dimensionality of these embeddings is reduced. 
Subsequently, the reduced embeddings are clustered into semantically similar groups, i.e., topics. The reduction of dimensionality is deliberately placed before clustering to increase computational efficiency as well as clustering accuracy \citep{allaoui2020}.
In the fourth step, topics are tokenized.
Finally, tokens are weighted.
To enhance the quality of extracted topic representations, \citet{grootendorst2022} introduces class-based tfidf (c-tfidf), which weighs the importance of tokens within topics, enabling a more efficient extraction of topic representations. 

Despite its significant advantages, BERTopic also faces several drawbacks.
It tends to produce a manifold of closely interconnected topics which may vary upon repeated modeling attempts \citep{egger2022}. 
This variability contributes to inconsistency in producing meaningful results, further complicated by the complexity of interpreting hyperparameters, hindering troubleshooting and diminishing the reliability of results \citep{abdelrazek2023}.
Moreover, BERTopic assumes that each document relates to a single topic, potentially oversimplifying real-world document complexity \citep{grootendorst2022}. 
Additionally, sentence-transformer models used for document embedding perform optimally with sentences or paragraphs \citep{reimers2019}. 
Furthermore, high computation times can result from processing large amounts of data \citep{grootendorst2022}.

Due to its novelty, applications and enhancements of BERTopic are still in their infancy. 
In a financial context, \citet{kim2022} utilized BERTopic on Item 1A from 10-K filings. They assessed whether identified topics can enhance the accuracy of ESG rating predictions and quantify each topic's relative contribution to the final rating prediction.
In other contexts, BERTopic has been applied in various studies: \citet{sanchez-franco2022} analyzed customer reviews, \citet{abuzayed2021} explored its application with pre-trained Arabic language models, \citet{egger2022} evaluated its performance on Twitter data, and \citet{grigore2023} extended BERTopic to predict individual's responses to a questionnaire based on their social media activity.

\subsection{Topic Modeling of Item 7 and Item 7A}

Our research is driven by several motivations regarding the choice of documents and analysis techniques.
Item 7 and Item 7A stand out as particularly crucial sections in 10-K reports \citep{bhattacharya2024}.
The MD\&A section (Item 7) provides a narrative that contextualizes the presented numbers, covering topics such as performance, liquidity, risks, and operations. 
In this section, management offers its individual perspective, which is essential for understanding the company's strategic direction and potential challenges. 
Additionally, the MD\&A section offers the most leeway and flexibility, making it rich with insights and indicative of future performance \citep{cohen2020}. 
Item 7A focuses on market risks, containing valuable information regarding the company's prospective performance. 
Hence, analyzing Item 7 and Item 7A allows us to uncover hidden textual information that has the potential to support the prediction of a company's future performance.
A technique that shows promise in addressing this challenge is topic modeling \citep{ranta2022}.
Despite advancements in computational capabilities and the emergence of new topic modeling techniques, there remains a gap in applying topic modeling methods to financial texts, particularly Item 7 and Item 7A. 
Furthermore, LDA continues to dominate applied topic modeling, although newer approaches like BERTopic offer potential improvements \citep{egger2021, blair2020}.

In this paper, we aim to demonstrate how FinTextSim, a finetuned sentence transformer, outperforms the most widely used sentence transformer AM on text from Item 7 and Item 7A from 10-K reports of S\&P 500 companies.
Additionally, we hypothesize that combining BERTopic with FinTextSim will significantly enhance the quality of financial information, providing better insights for stakeholders. 
We foresee that FinTextSim will also facilitate the application of aspect-based sentiment analysis, eventually improving business valuation and stock price prediction models.

\section{Materials and Methods}
In the following subsections, we outline the materials and methods of our study. This section is divided into several parts: sourcing the dataset, creating an enhanced financial keyword list, training FinTextSim, creating the topic models, and presenting the metrics used to evaluate the performance of the topic models.

\subsection{Dataset}
Our study focuses exclusively on Item 7 and Item 7A of 10-K reports while avoiding survivorship bias and ensuring the highest possible document comparability.
Given their greater significance, we deliberately choose 10-K over 10-Q reports \citep{griffin2003}.
We source our data from the Notre Dame Software Repository for Accounting and Finance in text-file format, which underwent a 'Stage One Parse' to remove all HTML tags.\footnote{The data can be found at: https://sraf.nd.edu/data/stage-one-10-x-parse-data/.}

To prevent survivorship bias, we filter 10-K filings of all companies that have been listed in the S\&P 500 index between 2015 and 2022. 
Using a regular expression-based extractor, we isolate the text from the start of Item 7 to the start of Item 8. Through this endeavor, we obtain the raw text of Item 7 and Item 7A. We refer to this combination of Item 7 and Item 7A as 'documents'. 
In order to maintain comparability, documents containing fewer than 250 words are discarded.\footnote{Paragraphs typically consist of 100–200 words. 
Moreover, sentence-transformers, such as AM and FinTextSim are designed to capture the semantic information of sentences and short paragraphs. 
Input texts longer than 256-word pieces (approximately 170-210 words) are truncated by default. 
The 250-word threshold ensures that each document includes at least two paragraphs, enhancing relevance, as shorter texts often lack substantive or complete ideas.}

Subsequently, we remove further outlier documents, identified by z-score. 
The z-score is a statistical measure that quantifies the distance of data points to the mean of a dataset, taking the standard deviation into account. We define data points as outliers if they deviate more than two standard deviations from the mean. Subsequently, we focus on documents from the period between 2016 and 2022.
%
Finally, we apply classical text processing techniques to identify relevant documents, ensuring that only meaningful texts are included in the analysis.
As part of this process, we normalize documents by removing stopwords, applying lemmatization, and performing tokenization. 
Subsequently, documents with low cosine similarity to others are filtered out.
This procedure ensures that both classical (evaluated in \ref{sec:results_classical_tq}) and contemporary models are applied to the same set of documents for a direct comparison.
The text fed into the sentence transformers remains unprocessed by classical techniques.
The number of documents retained at each preprocessing step is shown in Table \ref{Dataset.}.
\begin{table}[!ht]
\centering
\caption{Dataset.}
\label{Dataset.}
\begin{tabular}{lccc}
\toprule
 Preprocessing-Step & \# documents \\ 
\midrule
Documents extracted related to S\&P 500 companies & 4,600 \\
Documents with less than 250 words & 1,019 \\
Documents with z-score greater than 2 & 165 \\
Documents outside the timeframe of 2016-2022 & 373 \\
Documents with low cosine similarity & 1,604 \\
\midrule
Remaining documents in database & 1,439 \\
\bottomrule
\end{tabular}
\end{table}

\subsection{Keyword List}
\label{sec:keyword_list_explanation}
To train FinTextSim we enhance and utilize a keyword list based on the works of \citet{li2010a} and \citet{fengler2023}.
The economic anchorword list for 10-K and 10-Q reports encompasses 11 distinct topics: sales, cost, profit/loss, operations, liquidity, investment, financing, litigation, employment, tax/regulation, and accounting \citep{li2010a}. 
\citet{fengler2023} augmented it to a topic-word list by identifying semantically similar words, using a Word2Vec model trained on MD\&A sections of 10-K reports. 
We further enhance this list by adding key performance indicator names associated with company performance (e.g., EBITDA or EBIT) and supplementing the operations topic with terms related to logistics, supply chain and marketing. 
Additionally, we expand the topic-keyword list to 14 topics by introducing three topics that have garnered significant attention in recent years:
\begin{itemize}
\item Energy, in response to the 2022 energy crisis \citep{ferriani2023,canelli2024},
\item Environmental Sustainability, driven by the rising demand for ESG criteria \citep{cohen2023,gu2023,gu2024}, and
\item COVID-19 pandemic \citep{borrero2024,dechow2021,lin2021}.
\end{itemize}
An excerpt of the topics and the most important keywords is displayed in \ref{sec:keyword_list}.

\subsection{FinTextSim}
\label{sec:finsim}
To accurately cluster semantically similar financial text, we introduce FinTextSim, a sentence-transformer model specifically finetuned to enhance contextual embeddings for the financial domain.
Given the financial jargon and its domain-specific nuances, off-the-shelf (OTS), general-purpose sentence transformers fall short. 
Existing models tailored for the financial domain are primarily optimized for sentiment analysis (e.g. \citet{araci2019,LI2023163,GUO2024103371}). Hence, they also prove insufficient.
By fine-tuning FinTextSim on financial text, we aim to improve the quality of generated topics, enhancing semantic coherence and ensuring greater separation between topics. We hypothesize that this specialized fine-tuning will bridge the gap between general-purpose models and the specific demands of financial text analysis.

To tailor FinTextSim for financial text, we create a labeled dataset by adopting a dictionary-based approach, relying on the keyword list presented in Section \ref{sec:keyword_list_explanation}. 
We build upon our dataset and remove noise by replacing contractions as well as URLs and numerical characters.
After tokenizing the documents into sentences, we remove noisy ones by discarding those with less than five and more than 50 words.
To create a labeled dataset, we construct a keyword-sentence matrix. 
We iterate over each word in each sentence, checking if a sub-string matches a keyword. We deliberately use a sub-string approach to overcome issues with exact matches. For instance, our keyword list includes the word 'logistic'. Using sub-strings correctly allows words like 'logistics' and 'logistical' to be recognized as a keyword match.
Additionally, we make minor adjustments to the presented keyword list to prevent the double-counting of keywords. For example, to avoid 'cashflow' generating two entries for both 'cash' and 'cashflow', the refined list only includes 'cash'. 
After creating the keyword-sentence matrix, sentences containing two or more keywords from only one specific topic domain are labeled accordingly, ensuring topics are distinct. 
Moreover, we consider unique sentences to prevent overemphasis on particular wordings during training.

To address the initial imbalance in labeled sentence distribution and enhance FinTextSim’s generalization capabilities, we expand the dataset through multiple strategies. 
First, we incorporate an external dataset containing annual and sustainability reports from S\&P 500 companies\footnote{Dataset available at: https://huggingface.co/datasets/lemousehunter/SnP500-annual-and-sustainability-reports}, following the same keyword-based labeling logic. This additional dataset broadens the scope of financial discourse captured by FinTextSim, ensuring a more comprehensive representation of domain-specific language. 
Second, for the underrepresented topics of "Litigation" and "Covid Pandemic" in the original Item 7 dataset, we relax keyword-matching constraints to increase the number of labeled instances. This adjustment ensures that these critical topics receive adequate representation in the training process. 
Finally, we augment sentences for the least represented topic of "Litigation" using ChatGPT. Through this endeavor, we provide further balance and improve FinTextSim’s ability to accurately distinguish financial topics.

Following these steps, our dataset comprises 180,435 labeled sentences, split into training and test sets with an 80/20 ratio. 
To ensure balanced learning across all topics, we perform the test-train-split topic-wise.
Following these steps, we obtain 36,094 test- and 144,341 train-sentences.

To train FinTextSim, we employ adaptive circle loss and follow methods outlined by \citet{reimers2019,devlin2019}.
As base model, we select ModernBERT, a recent advancement in encoder-only architectures introduced by \citet{warner2024}. 
We adapt ModernBERT with a mean pooling and a normalization layer to enhance its performance for sentence similarity tasks \citep{reimers2019}.
We train the model using adaptive circle loss, deliberately choosing it over common triplet-based approaches.
While triplet loss minimizes the distance between similar samples and maximizes it for dissimilar ones, it applies equal penalty strength to positive and negative pairs.
In contrast, circle loss dynamically adjusts penalties based on how far a similarity score deviates from its optimal value, allowing more targeted optimization. 
It focuses on less optimized pairs while mildly adjusting those already close to convergence. 
Additionally, its circular decision boundary reduces ambiguity in feature space separability \citep{sun2020}.
Adaptive circle loss further extends this approach by employing curriculum learning, dynamically adjusting margin and scale to improve convergence and stability.
We train FinTextSim with a batch size of 200, an initial scale of 5, and an initial margin of 0.25, progressively increasing the scale to 16 and decreasing the margin to 0.1.

To evaluate performance, we encode the test dataset with AM and FinTextSim, comparing intra- and intertopic similarity (see \ref{sec:organizing_power}). 
AM, the most downloaded model for sentence similarity tasks on Hugging Face, serves as a robust baseline for evaluating out-of-the-box sentence-transformer embeddings. 
By benchmarking FinTextSim against AM, we assess whether specialized contextual embeddings improve financial text analysis.
We hypothesize that integrating domain-specific knowledge into contextual embeddings, will enable FinTextSim to generate more accurate and semantically meaningful representations of financial text than general-purpose models like AM. 
This improvement is crucial for precisely identifying meaningful financial topics and ensuring more reliable document organization.
To further examine the structure of the learned embeddings, we visualize them by reducing dimensionality to two dimensions with Uniform Manifold Approximation and Projection (UMAP).
Compared to alternatives, such as t-SNE or PCA, UMAP more effectively preserves both local and global structure \citep{allaoui2020, angelov2020}.
For UMAP, we employ the following essential hyperparameters:
\begin{itemize}
    \item Minimum distance: 0, to encourage closely grouped data points, facilitating the formation of clusters representing semantically similar documents.
    \item Distance metric: Cosine similarity, widely used in NLP and similarity tasks.
    \item n\_neighbors: 100, prioritizing global structures in our data to identify overarching macrotopics as well as hierarchically lower-ranked microtopics \citep{angelov2020}.
\end{itemize}

\subsection{Model Creation}

\label{sec:contemporary_approaches_creation}
Since BERTopic relies on sentence transformers optimized for sentence-level input and assumes that each document belongs to a single topic \citep{grootendorst2022}, we focus exclusively on sentence-level processing. 
To achieve this, we tokenize the documents displayed in Table \ref{Dataset.} into individual sentences. 
This approach is necessary as full MD\&A sections typically cover multiple topics. Treating entire documents as single-topic texts would yield misleading results.
Additionally, sentence-transformer models like AM and FinTextSim are designed for short texts, with a processing limit of 256-word pieces. 
Longer inputs are truncated, risking the loss of critical information. By working at the sentence level, we ensure compatibility with these models while preserving topic integrity.

Although BERTopic is known to leverage noise to create contextualized embeddings, we fit the models on sentence and refined sentence input.
Specifically, we refine the sentences by retaining those containing at least one keyword from our list as well as the pre- and succeeding sentences to maintain context and relevance \citep{altaweel2019}.
As a result, we obtained 687,959 sentences and 678,218 refined sentences.

We generate contextual embeddings for our dataset using FinTextSim and AM, applying each to BERTopic with identical models and hyperparameter settings to ensure that only the embedding choice influences performance.
In terms of dimensionality reduction, we opt for UMAP due to its ability to preserve both global and local structures \citep{allaoui2020, angelov2020} and its scalability to large datasets \citep{angelov2020}.
We specifically configure UMAP with the same settings as displayed in Section \ref{sec:finsim}. 
To strike a balance between clustering efficiency and information retention, we reduce the dimensionality to ten components.
For clustering, we choose Hierarchical Densitiy-Based Spatial Clustering of Applications with Noise (HDBSCAN) to address the limitations of centroid-based clustering algorithms in NLP and to effectively discern relevant from irrelevant data.
As a soft clustering algorithm, HDBSCAN models noise as outliers, ensuring that documents not fitting into any cluster are not forcibly assigned \citep{mcinnes2017}. 
Furthermore, HDBSCAN is proficient in detecting clusters of varying sizes and shapes, simplifies hyperparameter choices, and demonstrates computational scalability \citep{mcinnes2017}. 
Our specific choices for HDBSCAN's hyperparameters include:
\begin{itemize}
    \item Minimum Cluster Size: 1,250 to prioritize global topics over local ones.
    \item Minimum Number of Samples: 10, to reduce the number of outliers. The minimum number of samples determines the level of conservativity during clustering. If the number of samples is high, more data points are considered as noise, and the clustering is restricted to more dense areas.
\end{itemize}

We utilize a CountVectorizer for vectorization and exclude stopwords based on \citet{loughran2011} as well as infrequently occurring words. 
To extract relevant economic topics, we use c-tfidf weighting, reduce common words, and guide the model by incorporating seed words based on the keyword list with a multiplier of 50.

\subsection{Evaluation Metrics}
\label{sec:coherence}
To compare the performance of the topic models, we focus on two fundamental tasks \citep{blei2003}:
\begin{enumerate}
    \item Topic Quality: Uncover hidden topics in a collection of documents.
    \item Organizing Power: Organizing and structuring documents.
\end{enumerate}
The following subsections provide detail in how we objectively measure the model's capabilities for each task and their applicability for the financial domain.

\subsubsection{Topic Quality}
\label{sec:npmi_coherence}
We use NPMI coherence, known for its alignment with human judgment, to evaluate the quality of the generated topics. 
NPMI, a metric used in information theory and NLP, measures the strength of association between words by accounting for their individual frequencies. It is derived from Pointwise Mutual Information, which quantifies the difference between the actual co-occurrence of two terms and their expected co-occurrence if they were statistically independent.
NPMI coherence employs a sliding window approach to establish context vectors based on word co-occurrences \citep{roder2015}.
Given the typical sentence lengths in our dataset, we set the window size to 20, optimizing context coverage.\footnote{For classical topic modeling approaches, which are evaluated in \ref{sec:results_classical_tq}, we maintain the default window size of ten. However, due to the shorter sentence lengths resulting from stopword removal in classical models, we adjust the window size for BERTopic based on the ratio between sentence lengts of BERTopic versus classical models, guaranteeing comparable context coverage.}
Moreover, to mitigate vocabulary divergence effects, we lemmatize both the input texts and the extracted topic representations.
We use the five most representative words per topic as topic representations. 
While the standard procedure is to use ten representative words, using five to eight words is recommended to maintain clarity and conciseness \citep{agrawal2018}.

\subsubsection{Organizing Power}
\label{sec:organizing_power}
To assess the organizing power of FinTextSim and our topic models, we employ intra- and intertopic similarity.
To ensure clear, distinct and well-separated topic clusters, we aim to maximize intratopic similarity while simultaneously minimizing intertopic similarity.

We calculate intratopic similarity following this approach:
\begin{itemize}
    \item Topic Embedding Calculation: Compute the mean of all sentence embeddings for each topic to obtain topic embeddings.
    \item Cosine Similarity Calculation per Topic: Determine the cosine similarity of each sentence embedding within a topic to the corresponding topic embedding.
    \item Mean Calculation within Topics: The mean of these cosine similarities provides the intratopic similarity, representing the cohesion within a topic.
    \item Mean Calculation for all Topics: The mean of the intratopic similarities of each topic represents the model's intratopic similarity.
\end{itemize}

Intertopic similarity is computed as follows:
\begin{itemize}
    \item Topic Embedding Calculation: Compute the mean of all sentence embeddings for each topic to obtain topic embeddings.
    \item Cosine Similarity Matrix Computation: Calculate the cosine similarity between each topic embedding and all other topic embeddings, forming a pairwise similarity matrix.
    \item Extraction of Upper Triangle Values: Extract the upper triangle of the similarity matrix, excluding the self-similarity values in the diagonal.
    \item Mean Calculation for all Topics: The mean of these cosine similarities results in the model's intertopic similarity.
\end{itemize}
For the evaluation of FinTextSim, we use the topic labels from the test dataset. 
As we do not have any ground-truth labels for BERTopic, we employ its topic assignments.

\subsubsection{Applicability for the Financial Domain}
To evaluate the topic models specifically for the financial domain, we incorporate a topic-precision-based weighting into the metrics for topic quality and organizing power, using the previously described keyword list (see Section \ref{sec:keyword_list_explanation}).
Specifically, we weigh NPMI Coherence and intratopic similarity by multiplying them with topic-precision, while intertopic similarity is adjusted by dividing it by topic-precision.
We calculate topic-precision as follows:
\begin{itemize}
    \item Determine the dominant topic: A topic is classified as dominant if it contains at least two keywords from a single topic domain and no more than one keyword from another domain.
    \item Count True Positives (TP): The number of keywords belonging to the dominant topic’s original domain.
    \item Count False Positives (FP): The number of keywords from other topic domains. Keywords not present in the keyword list are ignored.
    \item Compute topic-precision: \begin{equation}
        Topic-Precision = \frac{TP}{(TP + FP)}
    \end{equation}
    \item Penalty for missing topics: If a topic from the keyword list is not captured, its topic-precision is set to 0.
    \item Assess overall performance: The model’s ability to capture financial topics is evaluated by averaging topic-precision across all topics.
\end{itemize}
This approach ensures that the detection of financial topics is the most crucial aspect of the topic models. 
If the identified topics lack financial relevance, their structural organization becomes obsolete. Moreover, if no financial topics are detected, the scores are penalized, reflecting the model’s unsuitability for financial text analysis.

\section{Results and Discussion}

We structure the results and discussion section according to our research questions:
\begin{itemize}
    \item[RQ1] FinTextSim: Leveraging the quality of contextual embeddings for the financial domain
    \item[RQ2] Topic Quality: Creating qualitative, coherent topic representations
    \item[RQ3] Organizing Power: Organizing large textual datasets
\end{itemize}
For FinTextSim, we highlight its result on the test dataset as well as its value in combination with BERTopic for topic modeling tasks. The results are presented and contextualized in the following subsections.

\subsection{FinTextSim - Leveraging Contextual Embeddings for the Financial Domain}
\label{sec:results_fintextsim}
\begin{figure}[!ht]
    \centering
    \subfloat[UMAP reduced sentence embeddings - FinTextSim. \label{fig:umap-reduced-sentence}]{
        \includegraphics[width=0.45\textwidth]{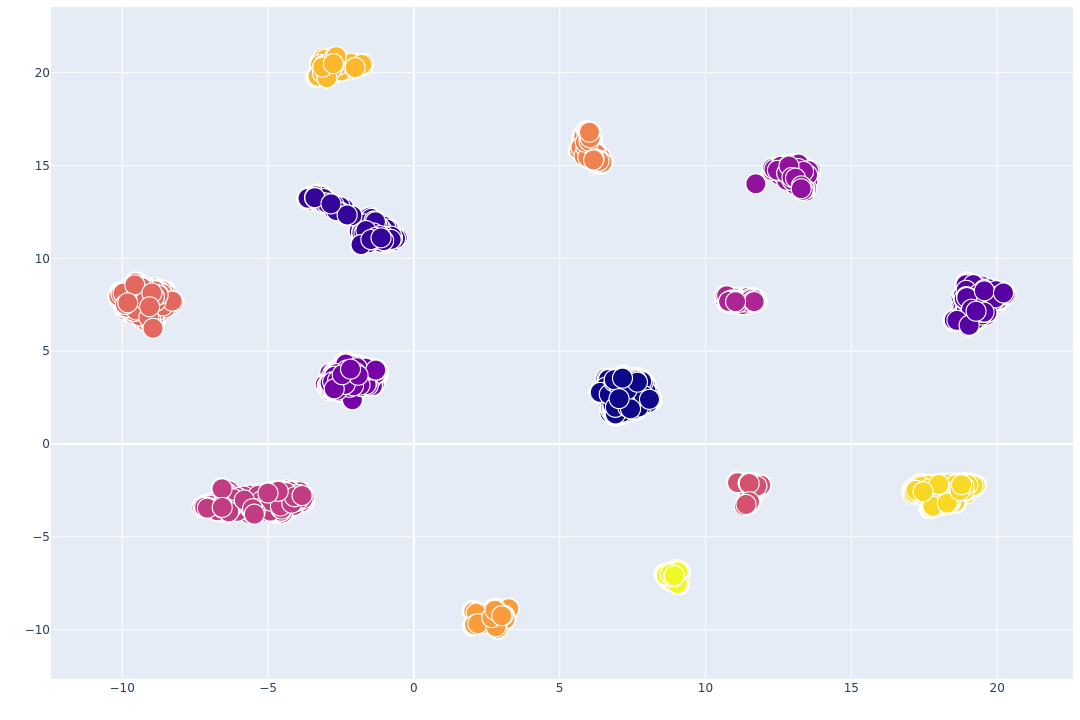}
    }
    \hfill
    \subfloat[UMAP reduced sentence embeddings - AM. \label{fig:second_plot}]{
        \includegraphics[width=0.45\textwidth]{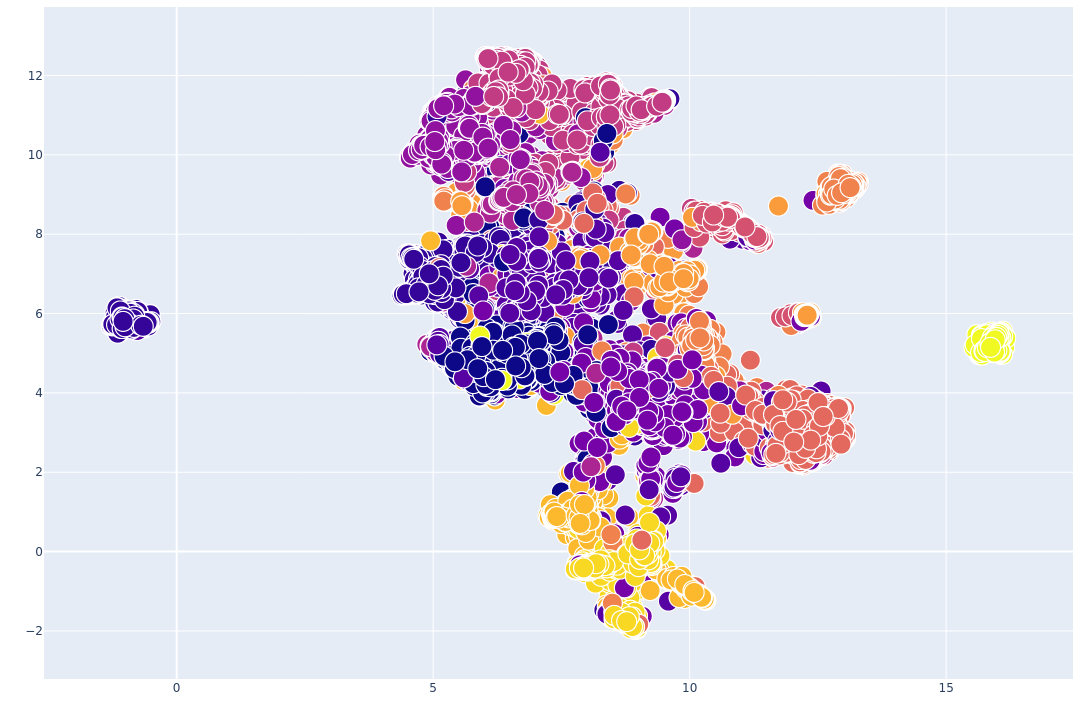}
    }
    \caption{FinTextSim vs. AM on the test dataset. \\ The colors of the datapoints represent a topic from the keyword list.}
    \label{fig:finsim_vs_ots}
\end{figure}

FinTextSim generates improved clusters and notably reduces the number of outliers in comparison to AM.
As illustrated in Figure~\ref{fig:finsim_vs_ots} and Table \ref{Intra- and Intertopic Similarity on test dataset}, FinTextSim leads to a significant increase of intratopic similarity, while simultaneously reducing intertopic similarity compared to AM on the test dataset. 
Specifically, FinTextSim increases intratopic similarity by 81\%, achieving a score of 0.9972, compared to 0.5498 for AM. 
Moreover, FinTextSim reduces intertopic similarity by 100\%, resulting in a score of 0.002, whereas AM has a score of 0.4647.
When combining BERTopic with AM, we identified 226,605 outliers. In contrast, FinTextSim generated 184,470 outliers, reducing the number of outliers by 19\%.
\begin{table}[!ht]
\centering
\caption{FinTextSim vs. AM: Intra- and Intertopic Similarity on test dataset.}
\label{Intra- and Intertopic Similarity on test dataset}
\begin{tabular}{lccc}
\toprule
 Model & Intratopic Similarity & Intertopic Similarity & Outliers within BERTopic \\ 
\midrule
FinTextSim & 0.9972 & 0.0002 & 184,470 \\
AM & 0.5498 & 0.4647 & 226,605 \\
\bottomrule
\end{tabular}
\end{table}

The results indicate that FinTextSim creates significantly better clusters of semantically similar concepts compared to AM.
FinTextSim's clusters are characterized by high intratopic similarity and low intertopic similarity.
In contrast, AM fails to accurately detect topic specifics, resulting in an indistinguishable mash of data points (see Figure~\ref{fig:finsim_vs_ots}).
This demonstrates that OTS sentence transformers are unsuitable for financial text. 

\begin{figure}[!ht]
    \centering
    \includegraphics[width=0.5\linewidth]{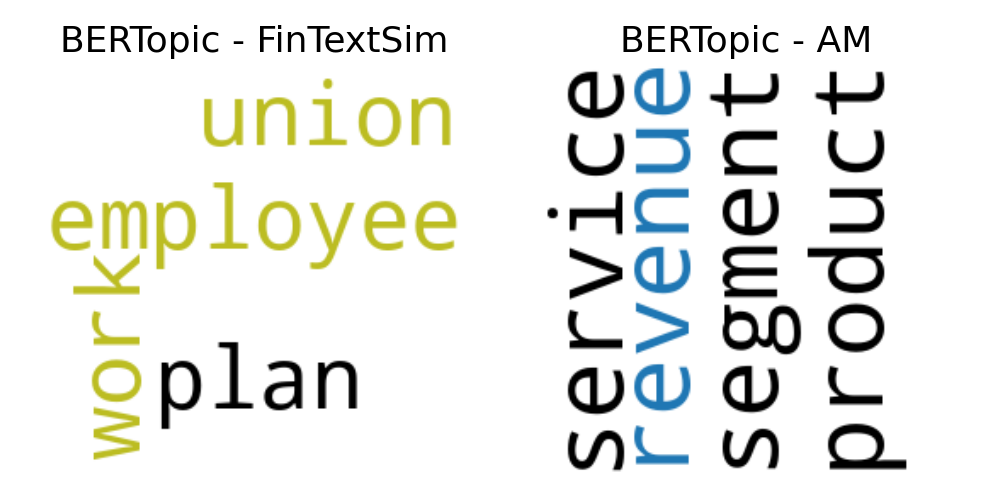}
    \caption{Topic representations - FinTextSim vs. AM - HR. 
    \\ Original cleaned sentence: 'a majority of employees belong to labor unions'.
    }
    \label{fig:topic_reps_bertopic_hr}
\end{figure}
Turning to a practical example, BERTopic with FinTextSim is able to accurately capture the underlying topic of a sentence, while the results with AM are misleading.
Figure~\ref{fig:topic_reps_bertopic_hr} displays word clouds for topics assigned to the same sentence using BERTopic with FinTextSim and AM.
FinTextSim correctly identifies that the sentence focuses on 'HR and Employment'. 
In contrast, AM fails to detect the topic, leading the analyst to associate it incorrectly with revenue and products.

\subsection{Topic Quality}
As displayed in Section~\ref{sec:coherence}, we use NPMI coherence weighted by topic-precision to objectively measure the topic quality of each topic model.
Table~\ref{Topic Precision Scores} displays the topic-precision scores for each topic model.
\begin{table}[!ht]
\centering
\caption{Topic-Precision Scores.}
\label{Topic Precision Scores}
\begin{tabular}{lcc}
\toprule
 & \multicolumn{2}{c}{Type} \\ \cmidrule{2-3}
 Model & Sentences & Refined Sentences \\ 
\midrule
AM & 0.311 & 0.684 \\
FinTextSim & 1.0 & 1.0 \\
\bottomrule
\end{tabular}
\end{table}

Using BERTopic with FinTextSim significantly enhances the identification of economic topics. 
For sentence-level input, AM achieves a topic-precision of 0.311 while failing to detect ten out of 14 economic topics. 
Refining sentence input improves AM’s topic-precision to 0.684, reducing the number of missed topics to six. 
However, even with refinement, AM struggles to identify key financial themes such as Operations, Liquidity and Solvency, and Investment.
In contrast, FinTextSim correctly identifies all 14 economic topics, achieving a perfect topic-precision of 1.0.\footnote{The wordclouds for each model are displayed in Figures~\ref{fig:wordcloud_fintextsim} to \ref{fig:wordcloud_am_rel}.}

Although refining sentence input improves AM’s topic precision, general-purpose embedding models are insufficient for comprehensive financial text analysis. 
AM models fail to capture the full scope of financial topics, limiting their practical application. 
In contrast, FinTextSim not only identifies all relevant topics but also minimizes conceptual overlap, ensuring clearer topic distinctions and more effective document organization. 
These findings underscore the necessity of domain-specific embeddings for financial text processing, as general-purpose models fail to capture the nuances of economic language.

\begin{table}[!ht]
\centering
\caption{Coherence Scores.}
\label{Coherence Scores}
\begin{tabular}{lcc}
\toprule
 & \multicolumn{2}{c}{Type} \\ \cmidrule{2-3}
 Model & Sentences & Refined Sentences \\ 
\midrule
AM & 0.106 (0.341) & 0.257 (0.376) \\
FinTextSim & 0.279 (0.279) & 0.301 (0.301) \\
\bottomrule
\end{tabular}
\end{table}
Table \ref{Coherence Scores} presents the coherence scores, with non-topic-precision weighted coherence shown in parentheses. 
When incorporating topic-precision weighting, we find that BERTopic generates higher-quality financial topics using FinTextSim compared to AM. 
For sentence-level input, FinTextSim achieves a topic-precision weighted coherence score of 0.279, surpassing AM's 0.106 by 163\%. 
With refined sentence input, coherence scores improve across both models.
For AM, this increase is primarily attributed to its increased topic-precision. 
However, FinTextSim still outperforms AM by 17\%.

Contrary to our expectations, BERTopic achieves higher coherence with AM than with FinTextSim when topic-precision-based weighting is not applied.
Although FinTextSim better captures the distinct structure of financial text, as reflected by its high topic-precision, it yields lower coherence scores.
However, this result is misleading in the context of financial text analysis. 
AM generates significantly more outliers and fails to capture key economic topics, leading to a loss of valuable information, potentially compromising studies. 
We suspect that the higher coherence of BERTopic with AM results from the increased number of outliers, which simplifies the compression and generation of topics.
Another factor is the vocabulary of the financial domain. Financial terms are often standalone words that do not necessarily co-occur within a sliding window. 
Therefore, coherence scores do not always align with human judgment of topic quality.\footnote{This trend is also observed for classical models (see \ref{sec:results_classical_tq}).}
Relying solely on standard coherence metrics without domain-specific weighting can be problematic. 
In financial text analysis, ensuring high topic-precision is crucial as the meaningful organization of domain-specific knowledge must take precedence over raw coherence scores.
Since topic evaluation requires domain expertise and subjective interpretation \citep{grootendorst2022}, standard coherence metrics alone are insufficient.
Without domain-specific weighting, coherence fails to reflect practical applicability, making topic-precision essential for capturing meaningful financial insights.

\begin{figure}[H]
    \centering
    \includegraphics[width=0.5\linewidth]{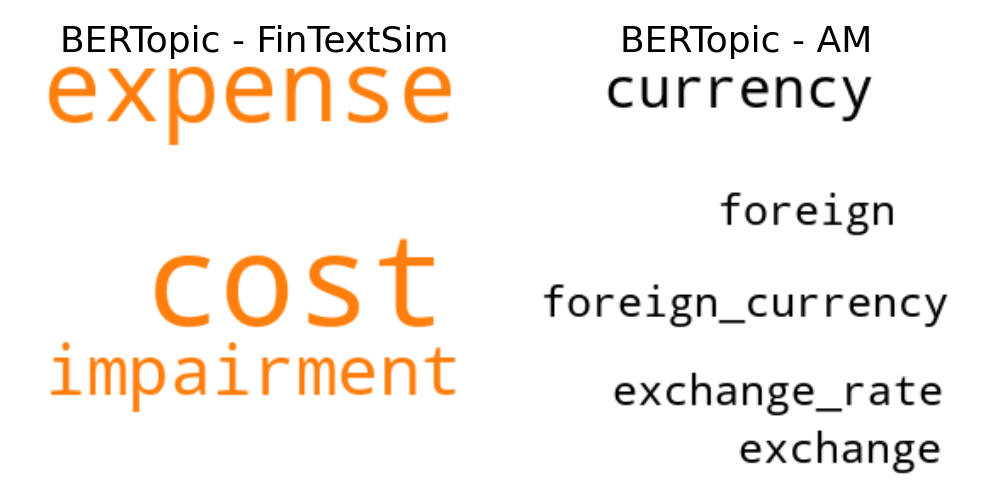}
    \caption{Topic representations - FinTextSim vs. AM - Cost. 
    \\ Original cleaned sentence: 'business is vulnerable to fluctuations in fuel costs and disruptions in fuel supplies'. 
    }
    \label{fig:topic_reps_bertopic_cost}
\end{figure}
To illustrate this in a practical example, we refer to Figure~\ref{fig:topic_reps_bertopic_cost}.
Once again, FinTextSim correctly identifies the underlying 'Cost' topic, while AM misclassifies it, associating it with currency and exchange rates.
Despite this misclassification, AM receives a higher coherence score of 0.448 compared to FinTextSim's 0.324.

\begin{figure}[H]
    \centering
    \includegraphics[width=0.5\linewidth]{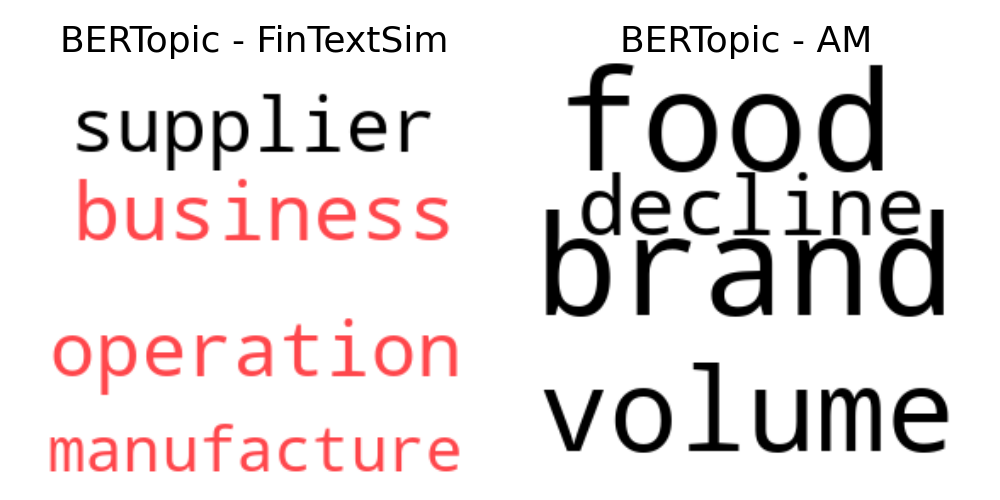}
    \caption{Topic representations - FinTextSim vs. AM - Operations. 
    \\ Original cleaned sentence: 'the company manufactures markets and distributes spices seasoning mixes condiments and other flavorful products to the entire food industry retailers food manufacturers and foodservice businesses'. 
    }
    \label{fig:topic_reps_bertopic_operations}
\end{figure}
For further clarity, Figure~\ref{fig:topic_reps_bertopic_operations} presents another practical example.
FinTextSim correctly identifies the 'Operations' topic, whereas AM misclassifies it as a non-economic concept.
Yet, FinTextSim receives a lower coherence score of 0.205 compared to AM's 0.262.
These discrepancies underscore the limitations of coherence scores in evaluating financial text, as they fail to account for domain relevance and topic-precision.

\subsection{Organizing Power}
\label{sec:results_op}

To efficiently organize and structure large collections of documents, maximizing intratopic similarity while simultaneously minimizing intertopic similarity is desirable. 
The results for intratopic similarity of our models are displayed in Table \ref{Intratopic Similarity}. Non-topic-precision weighted scores are illustrated in parentheses.
\begin{table}[!ht]
\centering
\caption{Intratopic Similarity.}
\label{Intratopic Similarity}
\begin{tabular}{lcc}
\toprule
 & \multicolumn{2}{c}{Type} \\ \cmidrule{2-3}
 Model & Sentences & Refined Sentences \\ 
\midrule
AM & 0.206 (0.661) & 0.441 (0.644) \\
FinTextSim & 0.925 (0.925) & 0.929 (0.929) \\
\bottomrule
\end{tabular}
\end{table}

FinTextSim consistently achieves higher intratopic similarity than AM, regardless of input type or the application of topic-precision weighting. 
When topic-precision weighting is considered, FinTextSim outperforms AM by a wide margin, achieving an intratopic similarity of 0.925 compared to 0.206 for sentence input, representing a 350\% improvement. 
For refined sentence input, FinTextSim reaches 0.929, outperforming AM’s 0.441 by 111\%. 
Even without topic-precision weighting, FinTextSim maintains a strong advantage, with a 40\% increase for sentence input, scoring 0.925 compared to AM’s 0.661. 
The trend remains consistent for refined sentence input, where FinTextSim achieves 0.929, exceeding AM’s 0.644 by 44\%. 
These results further highlight FinTextSim’s ability to generate more cohesive topic clusters, reinforcing its suitability for financial text analysis.

The intertopic similarities of the models are displayed in Table \ref{Intertopic Similarity}. Non-topic-precision weighted scores are illustrated in parentheses.
A good separation of the generated topics is represented by low intertopic similarities.
\begin{table}[!ht]
\centering
\caption{Intertopic Similarity.}
\label{Intertopic Similarity}
\begin{tabular}{lcc}
\toprule
 & \multicolumn{2}{c}{Type} \\ \cmidrule{2-3}
 Model & Sentences & Refined Sentences \\ 
\midrule
AM & 1.315 (0.409) & 0.631 (0.432) \\
FinTextSim & 0.064 (0.064) & 0.066 (0.066) \\
\bottomrule
\end{tabular}
\end{table}

FinTextSim significantly reduces intertopic similarity compared to AM, indicating enhanced topic separation. 
When topic-precision weighting is applied, FinTextSim achieves a 95\% reduction in intertopic similarity, scoring 0.064 compared to AM’s 1.315. 
With refined sentence input, intertopic similarity remains 90\% lower, with FinTextSim at 0.066 compared to AM’s 0.631. 
Neglecting topic-precision weighting, FinTextSim continues to outperform AM by a wide margin, reducing intertopic similarity by 84\% for sentence input, with scores of 0.064 versus 0.409. 
For refined sentence input, FinTextSim outperforms AM by 85\%, where it achieves 0.066 compared to AM’s 0.432. 
These results underscore FinTextSim’s effectiveness in minimizing topic overlap, leading to clearer and more distinct financial topic clusters.

\begin{figure}[!ht]
    \centering
    \includegraphics[width=0.5\linewidth]{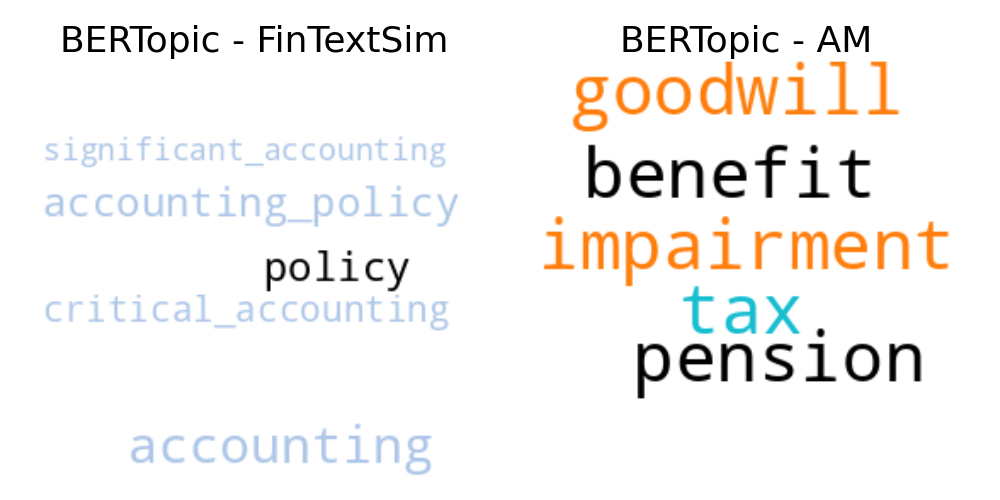}
    \caption{Topic representations - FinTextSim vs. AM - Accounting. 
    \\ Original cleaned sentence: 'critical accounting policies estimates and judgments our consolidated financial statements are based on gaap which requires us to make estimates and assumptions about future events that affect the amounts reported in our consolidated financial statements'. 
    }
    \label{fig:topic_reps_bertopic_accounting}
\end{figure}

Figure~\ref{fig:topic_reps_bertopic_accounting} illustrates this situation.
In this practical case, BERTopic with FinTextSim can accurately capture the underlying topic of the sentence, while the results with AM are misleading.
FinTextSim correctly recognizes that the sentence pertains to 'Accounting,' ensuring a precise and domain-relevant topic assignment. 
In contrast, AM fails to detect the specific topic, causing the analyst to misinterpret the sentence as relating to multiple themes, such as cost, tax, and HR. 
This inability to distinguish between economic topics results in high intertopic similarity and low intratopic similarity, highlighting the limitations of AM for financial text analysis.

\subsection{Wrapup of Results and Discussion}
We find that BERTopic is highly on financial text when combined with FinTextSim. 
AM, on the other hand, generates more general topics and fails to capture economic topics, leading to significant gaps in coverage. 
The quality of topics improves significantly when FinTextSim is used.
Only in combination with FinTextSim, BERTopic produce clear, distinct clusters of economic topics. 
In contrast, AM leads to frequent misclassifications.

Our findings support the hypothesis of \citet{dong2024}, demonstrating that a fine-tuned model grounded in a specialized dataset significantly improves both performance and domain-specific understanding. 
Furthermore, as \citet{gu2024a} indicates, finetuning a foundational base model enhances performance on complex tasks. 
Relying solely on OTS models may compromise reliability and introduce systematic errors, highlighting the importance of integrating fine-tuned models like FinTextSim for extracting meaningful and reliable insights.
However, the extent to which FinTextSim generalizes beyond 10-K reports remains an open question.
A small-scale experiment on Item 1 yielded similar results as described in Section \ref{sec:results_fintextsim}, suggesting that FinTextSim's effectiveness extends to other sections of 10-K filings.
Expanding its training data to include diverse financial sources, such as news articles, conference call transcripts, and analyst reports, could further enhance its generalization capabilities.
Additionally, incorporating researcher-labeled data may provide further improvements in FinTextSim’s adaptability and robustness across financial contexts.

Regarding our results, it is important to note that the displayed metrics should never be considered in isolation, especially those regarding organizing power. 
For instance, even if AM would show high intratopic and low intertopic similarity, it does not necessarily produce 'good' clusters, as the quality of the generated topics may remain low. 
In such cases, its ability to enhance organizational clarity would still be limited. 
Therefore, evaluating topics requires looking beyond the raw metrics to consider their true quality.

Hence, evaluating topic models remains challenging \citep{zhao2021}.
Our analysis reveals the limitations of coherence as a measure. For instance, BERTopic with AM achieves higher coherence scores than FinTextSim. 
Yet, we identified low topic-precision scores, indicating numerous missing economic topics and/or overlapping concepts. This suggests that higher coherence does not necessarily correlate with higher topic quality. 
Our findings underscore the necessity for new coherence or topic quality measures, particularly for domain-specific texts like finance. In such texts, topic words often stand alone and may not co-occur within a sliding window. 
Hence, traditional coherence metrics cannot capture the 'true' quality of the generated topics.

While BERTopic enhances topic modeling compared to the classical approaches, there is still significant room for improvement. 
The transformer architecture, which BERTopic heavily relies on, may not be fully optimized yet. Thus, more sophisticated and computationally efficient alternatives should be explored \citep{karami2024}.
Further advancements in encoder-only models could enhance sentence transformers by improving their contextual understanding of language \citep{warner2024}. 
Moreover, applying domain-specific pre-training methods to optimized BERT variants may deepen the model’s understanding of financial language, leading to more effective downstream task performance \citep{huang2023}.
Additionally, BERTopic's inconsistency in producing meaningful results, compounded by the complex hyperparameters of the underlying models, compromises reliability \citep{abdelrazek2023}.
Hence, future research should focus on developing an objective standard for selecting models and tuning hyperparameters. Specifically, we plan to investigate the impact of hyperparameter tuning for both dimensionality reduction and clustering techniques on contextual embeddings. This approach aims to streamline the process of topic modeling, objectively determining hyperparameters. Eventually, this will improve topic clustering and extraction, thus enhancing the analysis of textual data.

\section{Conclusion}

Increased availability of information and enhanced computational capabilities have transformed the analysis of annual reports, recognizing the value embedded within qualitative textual data. 
Automated review processes, such as topic modeling, are crucial for analyzing this data.
However, in our domain, the use of those ML-based methods, including contextual embeddings, remains underexplored \citep{ranta2022}.
We address these issues by introducing FinTextSim, a finetuned sentence transformer enhancing analysis of financial text with BERTopic.

Our study reveals the significant advantages of FinTextSim over OTS sentence-transformer models.
FinTextSim excels in generating distinct clusters of topics, substantially outperforming OTS sentence-transformer models on financial text.
This highlights the need for domain-specifically finetuned sentence-transformer models.
Additionally, FinTextSim allows BERTopic to identify the most qualitative topics for the financial domain. While FinTextSim captures all relevant financial topics, OTS sentence-transformers miss valuable information. 
Combining BERTopic with FinTextSim notably enhances its capability to create well-separated clusters of economic topics. Hence, the domain-specific fine-tuning of sentence transformers is crucial for achieving optimal topic modeling outcomes.

Through our efforts, we make several significant contributions.
First, we enhance contextual embeddings for our domain with FinTextSim, amplifying results and insights from future studies.
Moreover, FinTextSim improves the quality of financial information, empowering stakeholders to gain a competitive advantage through more efficient allocation of resources and improved decision-making processes.
Furthermore, integrating FinTextSim into business valuation and stock price prediction models promises enhancements in accuracy and effectiveness, providing valuable tools for financial analysts and investors alike.

It is important to note that the evaluation of topic models remains a challenging task. 
Assessing each model's performance requires considering all presented metrics simultaneously, as relying on one measure in isolation may be misleading. Moreover, a qualitative analysis of topic representations is necessary.

Although BERTopic outperforms classical approaches, significant room for advancements in the field of topic modeling and their applications in the financial domain remains. 
Interesting directions for future research include the creation of an improved metric to objectively measure the quality of generated topic representations, particularly for domain-specific text like finance.
Additionally, advancing the architecture of sentence transformers or exploring new embedding techniques holds promise for further enhancement of topic modeling.
Moreover, there is a need for streamlining processes to determine optimal models and hyperparameters within BERTopic. 
Through this endeavor, topic instability will be addressed, enhancing the reliability and validity of results.
Integrating contemporary topic modeling approaches like BERTopic with FinTextSim could offer substantial benefits to new and ongoing studies, leveraging their results. 
Finally, harnessing the improved topic extraction and organizational structure generated with BERTopic in combination with FinTextSim holds the potential to significantly enhance business valuation and stock price prediction models, providing valuable insights for financial analysts and investors.

\newpage
\bibliography{references}

\newpage

\appendix
\section{Wordclouds BERTopic models}
\begin{figure}[H]
\centering
\includegraphics[width=\textwidth]{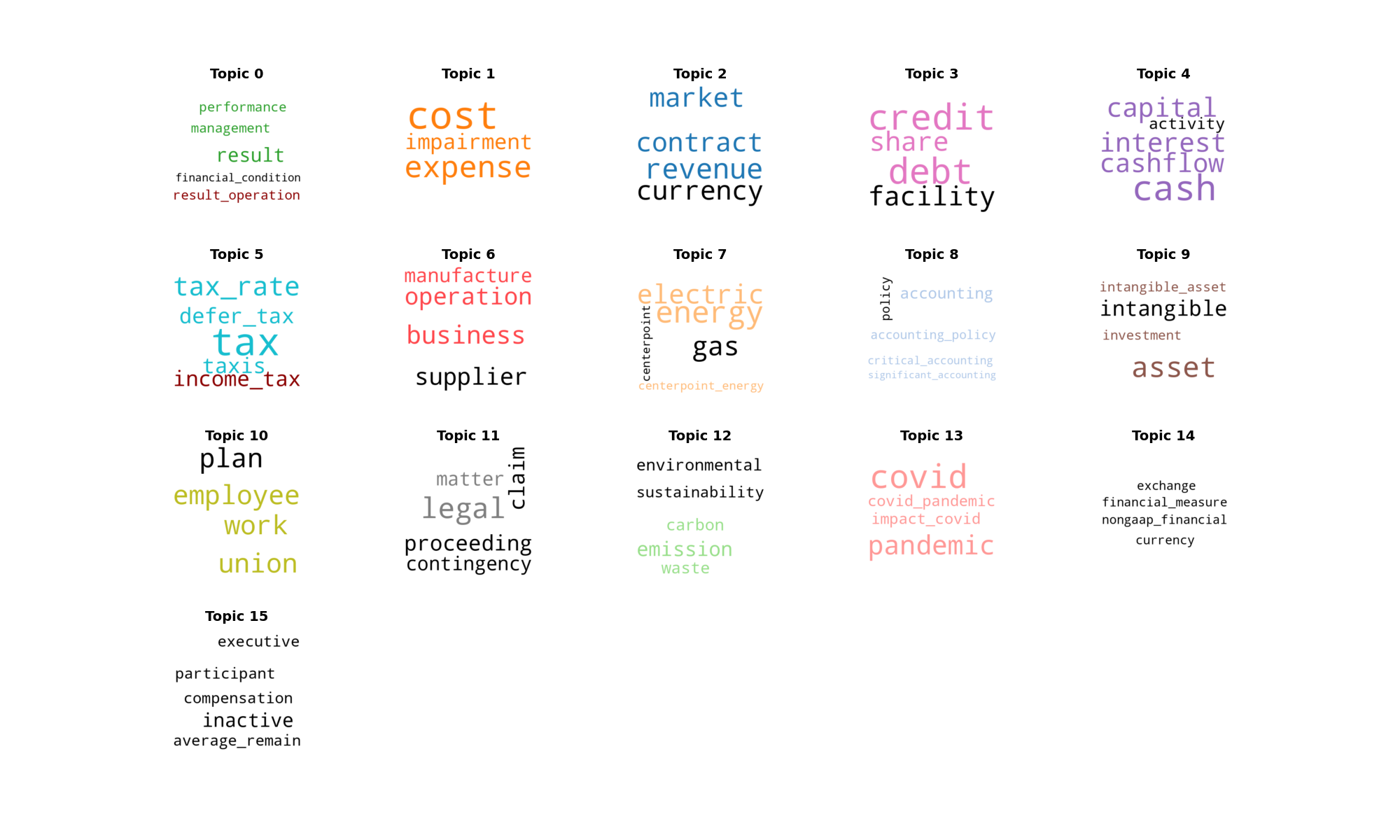}
\caption{Wordcloud - FinTextSim - Sentences.\\ The color of each word represents its associated unique topic from the keyword list. Words colored in black are not present in the keyword list. Words colored in darkred are bigrams containing words from multiple keyword domains.}
\label{fig:wordcloud_fintextsim}
\end{figure}

\begin{figure}[H]
\centering
\includegraphics[width=\textwidth]{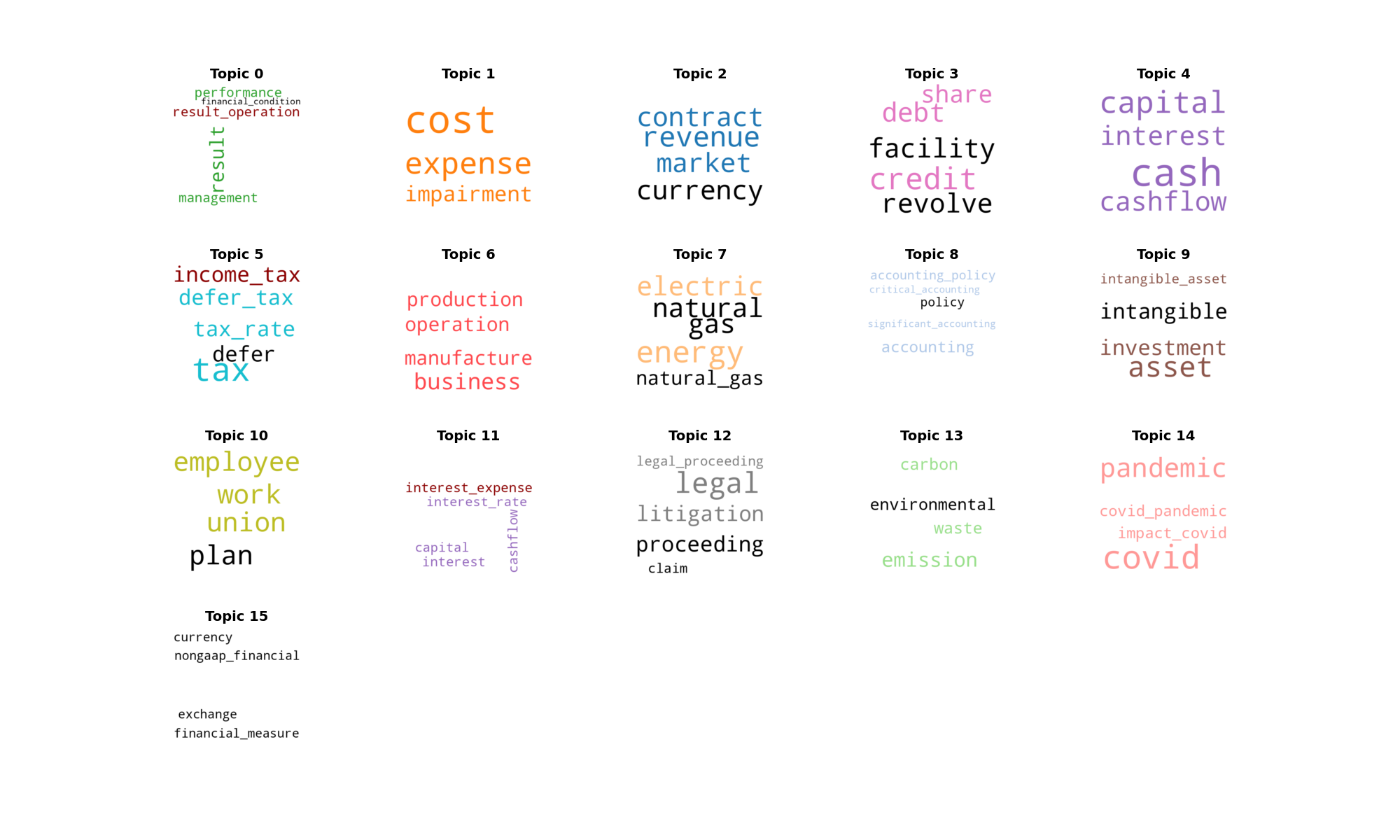}
\caption{Wordcloud - FinTextSim - Refined Sentences.\\ The color of each word represents its associated unique topic from the keyword list. Words colored in black are not present in the keyword list. Words colored in darkred are bigrams containing words from multiple keyword domains.}
\label{fig:wordcloud_fintextsim_rel}
\end{figure}

\begin{figure}[H]
\centering
\includegraphics[width=\textwidth]{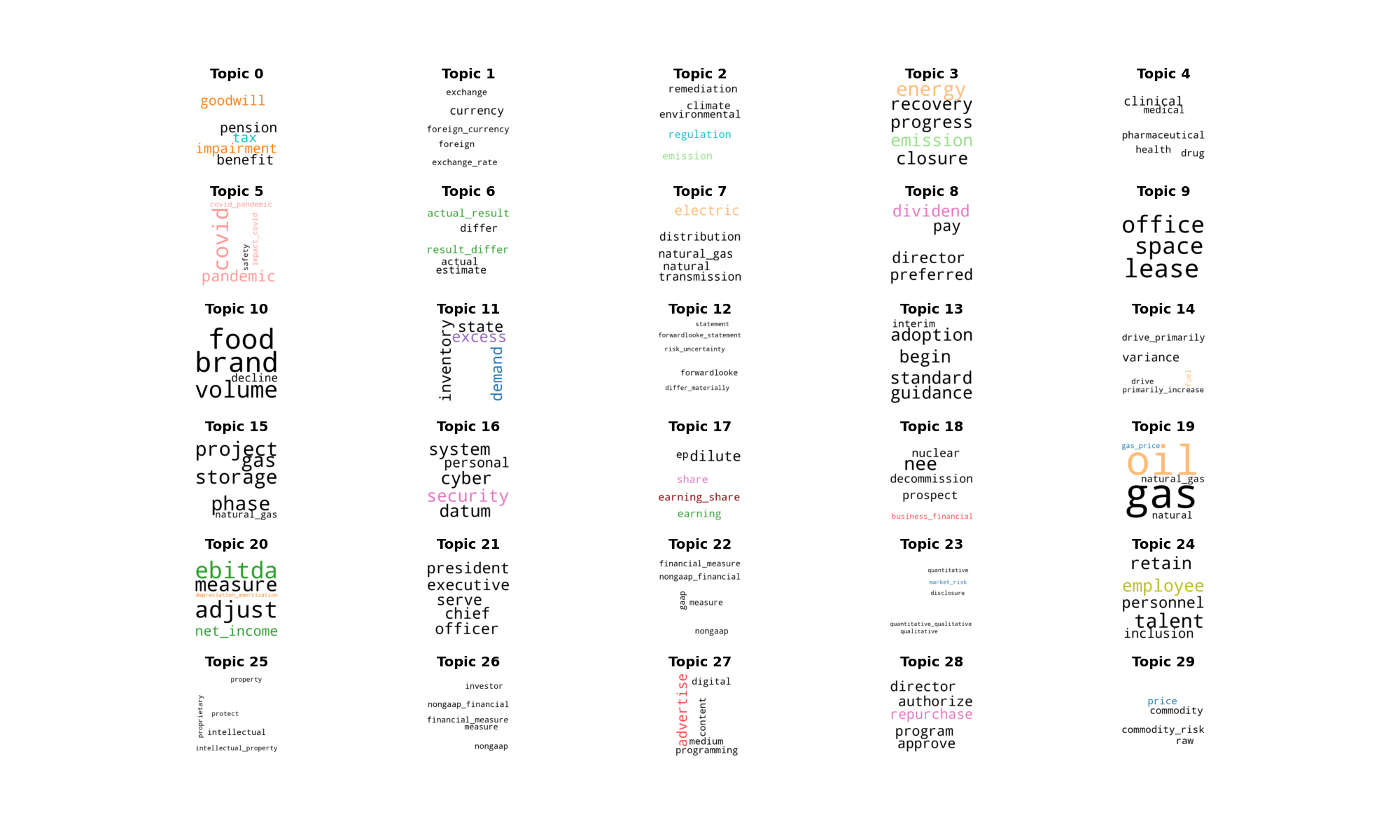}
\caption{Wordcloud - AM - Sentences.\\ The color of each word represents its associated unique topic from the keyword list. Words colored in black are not present in the keyword list. Words colored in darkred are bigrams containing words from multiple keyword domains.}
\label{fig:wordcloud_am}
\end{figure}

\begin{figure}[H]
\centering
\includegraphics[width=\textwidth]{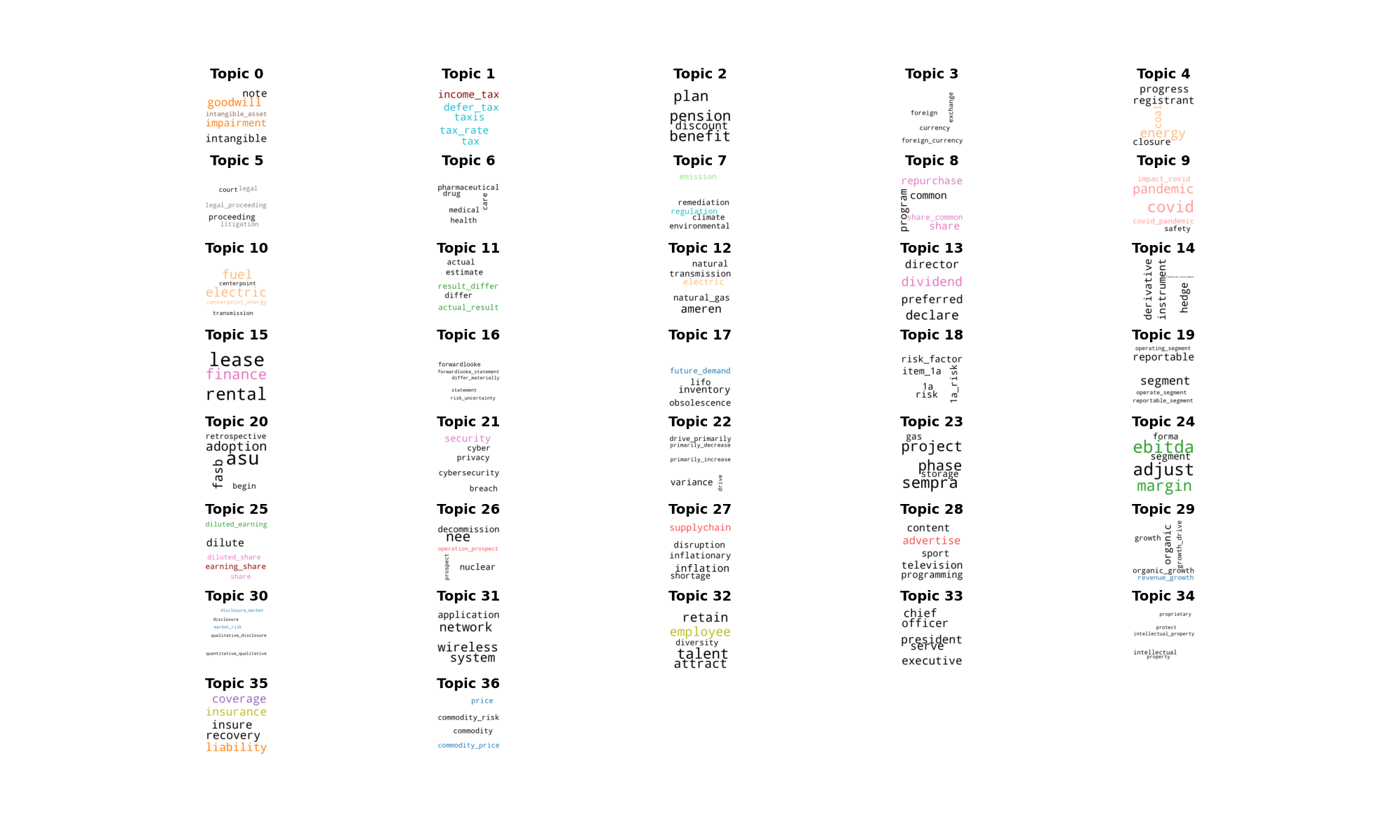}
\caption{Wordcloud - FinTextSim - Refined Sentences.\\ The color of each word represents its associated unique topic from the keyword list. Words colored in black are not present in the keyword list. Words colored in darkred are bigrams containing words from multiple keyword domains.}
\label{fig:wordcloud_am_rel}
\end{figure}

\section{Algorithms - Classical Topic Modeling Approaches}
For a deeper understanding of classical topic modeling approaches, we highlight two Bayesian models: LDA and Hierarchical Dirichlet Process (HDP), and one algebraic model: Non-Negative Matrix Factorization (NMF). 
The Bayesian models define a hypothetical generative process for documents, then work backwards to infer the topics that could have generated the documents \citep{abdelrazek2023}. 
In contrast, NMF factorizes a document-term matrix into term-topic and topic-document matrices \citep{lee1999}.
All three models operate under the BoW assumption, viewing each document as a mixture of underlying topics and each topic as a mixture of words. Hence, the algorithms assign prevalence of terms to topics ($\beta$) and topics to documents ($\gamma$) \citep{blei2003, teh2006}.
To ensure and enhance the performance of the presented classical models, several preprocessing steps are essential. These include tokenizing documents, removing stopwords, and lemmatization or stemming of words \citep{bellstam2021, fu2021, albalawi2020}.
The following subsections provide an overview of classical topic modeling techniques. 

\subsection{Latent Dirichlet Allocation}
The most prevalent topic modeling approach in literature is LDA.
LDA is a three-level parametric hierarchical Bayesian model, relying mainly on three hyperparameters \citep{blei2003}: the number of topics ($k$), the concentration parameter of the Dirichlet prior of the document-topic distribution ($\alpha$), and the parameter controlling the distribution of words across topics ($\eta$) \citep{fernandes2020}.
Each hyperparameter significantly influences the quality and stability of the generated topics. Yet, their selection remains challenging due to the inherent complexity of textual data \citep{maier2018, agrawal2018}.

Despite its widespread use, LDA has notable limitations. 
\citet{agrawal2018} discovered that LDA is sensitive to the order of training data, i.e., the topics vary when the training data is shuffled. Hence, systematic errors are incorporated into studies. By tuning the model's hyperparameters, these effects can be mitigated \citep{agrawal2018}.
Moreover, since LDA extracts topics from word distributions independently, overlapping topics can occur \citep{campbell2015}.

LDA has been used in various fields.
\citet{bao2014} pioneered the integration of unsupervised learning methods into Management Accounting and Finance using LDA to analyze risk disclosures from 10-K reports. 
\citet{dyer2017} examined topics contributing to the lengthening of 10-K reports over time, while \citet{brown2020} identified topics predicting financial misreporting. 
In additional financial research studies, LDA has been used to quantify the economic content in communications, identify central subjects, or estimate innovation capabilities, among other applications \citep{jegadeesh2017,lowry2020,bellstam2021,garcia2023,GAO2025103923}.

\subsection{Hierarchical Dirichlet Process}
HDP offers the flexibility of inferring the number of topics from the data itself, accommodating potentially infinite topics. Hence, this non-parametric hierarchical Bayesian model is particularly advantageous when the number of topics is uncertain \citep{teh2006}. 
However, the non-parametric nature of HDP increases the computational complexity \citep{fu2021,fernandes2020}.
Additionally, hyperparameter tuning is essential to enhance the quality and stability of topics. Each hyperparameter significantly influences the number of inferred topics, the distribution of words over topics, and the distribution of topics over documents \citep{fernandes2020}.

Compared to LDA, HDP is much less populated in literature, especially in terms of the financial domain. 
In a financial context, \citet{hong2018} used HDP to explore the evolution of topics of Chinese financial news. 
Outside finance, HDP has been used for content analysis \citep{altaweel2019}, text clustering \citep{fernandes2020, mayasari2021,yau2014} and to identify pertinent topics and satisfaction between different customer groups \citep{farzadnia2024}.

\subsection{Non-Negative Matrix Factorization}
NMF offers a decompositional, non-probabilistic approach to topic modeling, factorizing the input document-term-matrix $A$ into the product of term-topic-matrix $W$ and topic-document-matrix $H$ \citep{lee1999}.
By evaluating the discrepancy between $A$ and $W\times H$ using the squared Frobenius norm, the topic modeling problem is framed as an optimization task restricted to non-negative entries \citep{wang2023}. 
Unlike LDA and HDP, NMF does not rely heavily on hyperparameters, although the number of topics still needs to be specified by the user.

While NMF offers advantages in simplicity and computational efficiency \citep{egger2022}, it also faces several challenges.
Compared to LDA and HDP, it lacks a solid statistical foundation and a defined generative model.
Additionally, NMF relies on anchor words to enforce a block diagonal structure in the term-topic matrix $W$, ensuring consistent solutions \citep{donoho2003,gillis2014}. This assumption posits that each topic is associated with a unique anchor word, absent in other topics \citep{gillis2014}. Given the multifaceted nature of words, this assumption can be considered as fragile \citep{wang2023}.
Another assumption of NMF is that each topic has at least one 'pure document', a document discussing only that specific topic \citep{gillis2014}. This assumption is particularly fragile for longer documents.

NMF has applications in various fields and domains.
In finance, \citet{chen2017} used NMF and other topic modeling methods on 10-K and 8-K filings of bank holding companies to distinguish failed from non-failed banks. Additionally, \citet{cai2022} applied NMF to assess the impact of risk factor disclosures on bond pricing. 
In other fields, NMF has been primarily employed for short-text topic modeling \citep{chen2019, albalawi2020, egger2022}.

\subsection{Wrapup of Classical Topic Modeling Approaches}
Classical topic modeling approaches offer both advantages and disadvantages.
A main advantage is the easier interpretation of hyperparameters, aiding in troubleshooting and model interpretation. However, as model complexity increases, there is a growing risk of encountering inference problems \citep{abdelrazek2023}. 
Moreover, these models are particularly susceptible to the following issues: 
\begin{enumerate}
    \item BoW Assumption; context and semantic relationships cannot be captured \citep{murphy2024}; misrepresentation of topics and documents possible \citep{grootendorst2022},
    \item Interpretability of Topics \citep{campbell2015},
    \item Reliability, Validity, and Subjectivity: need for manual effort in preprocessing and hyperparameter selection \citep{baden2022}.
\end{enumerate}

\section{Model Creation - Classical Topic Modeling Approaches}
\label{sec:classical_approaches_creation}
For classical models, we assess their performance for short and long text modeling, as well as their capability of handling noise.
To achieve this, we utilize three input variants: one based on the entire documents, another based on individual sentences extracted from the documents, and a third based on further refined sentences using a keyword approach.
Specifically, we refine the sentences by retaining those containing at least one keyword from our list as well as the pre- and succeeding sentences to maintain context and relevance \citep{altaweel2019}.
Classical models require additional preprocessing steps.

Our further preprocessing steps, follow commonly adopted approaches, such as stopword removal, lemmatization and tokenizing.
We initiate by removing stop words using specialized lists tailored for financial applications provided by the Notre Dame Software Repository for Accounting and Finance\footnote{The stopword lists can be found at: https://sraf.nd.edu/textual-analysis/stopwords/}. 
Additionally, we maintain only words with specified part-of-speech tags\footnote{The part-of-speech tags to maintain are Nouns, Adjectives, Verbs, and Adverbs.}.
Subsequently, we refine our corpus and reduce vocabulary size by lemmatizing words. 
We deliberately choose lemmatization over stemming, as it preserves the interpretability of words better \citep{maier2018}.
Next, we tokenize the text into words to get the first draft of our tokenized documents. 
For sentence-based models, we add two additional steps. 
First, we segment the original documents into individual sentences. 
Then, we apply the previously mentioned steps. Additionally, we remove sentences that contain fewer than three or more than 50 words after these steps.

After obtaining the tokenized documents, we refine our dataset by enhancing document representations and removing noise in form of irrelevant terms and untypical documents.
First, we generate phrases and construct bigrams and trigrams, combining terms that require simultaneous analysis. 
Subsequently, we eliminate tokens that are either too infrequent or overly common across documents as well as those with low normalized tfidf-score
\footnote{For whole-document input, words must appear in at least 2\% and at most 99\% of the documents. For sentence-based input, the thresholds are 0.005\% and 15\% due to the shorter and more specific nature of sentences. The threshold for the normalized tf-idf score is set to 0.1 for all inputs.}. 
Throughout these steps, we ensure that terms from our keyword list are preserved. 
We then assess the total and unique token count per document, removing those documents with fewer than two tokens and/or two unique tokens. This step is particularly crucial for sentence-based models.
Finally, documents with a mean cosine similarity of less than 0.6 compared to all documents are discarded
\footnote{For sentence-based models, we discard documents with low cosine similarity based on the entire document. This ensures comparability of input and avoids computational inefficiencies.}. 
We deliberately remove low-similarity documents after completing all other preprocessing steps to improve document comparability before removal. The evolution of the number of documents and unique words per preprocessing step is displayed in Table \ref{Data Preprocessing Classical Models.}.
\begin{table}[!ht]
\centering
\caption{Data Preprocessing Classical Models.}
\label{Data Preprocessing Classical Models.}
\begin{tabular}{lccc}
\toprule
 & \multicolumn{3}{c}{Type} \\ \cmidrule{2-4}
 Preprocessing-Step & Whole document & Sentences & Refined Sentences \\ 
\midrule
\multicolumn{4}{l}{\textbf{Relevant Year Extraction}} \\
\# documents & 3,043 & 1,469,265 & 1,442,355 \\
\# words & 29,931 & 36,435 & 35,640 \\
\midrule
\multicolumn{4}{l}{\textbf{Phrases - Bigrams - Trigrams}} \\
\# documents & 3,043 & 1,469,265 & 1,442,355 \\
\# words & 30,019 & 36,706 & 35,907 \\
\midrule
\multicolumn{4}{l}{\textbf{Filter Extremes (tf and tfidf)}} \\
\# documents & 3,043 & 1,469,265 & 1,442,355 \\
\# words & 3,869 & 5,822 & 5,766 \\
\midrule
\multicolumn{4}{l}{\textbf{Few (unique) Tokens}} \\
\# documents & 3,043 & 1,462,490 & 1,435,988 \\
\# words & 3,869 & 5,780 & 5,766 \\
\midrule
\multicolumn{4}{l}{\textbf{Cosine Similarity}} \\
\# documents & 1,439 & 687,959 & 678,218 \\
\# words & 3,857 & 5,789 & 5,766 \\
\bottomrule
\end{tabular}
\end{table}

Upon completing the described preprocessing steps, we successfully reduce the number of documents and unique words. 
Specifically, the number of documents reduces from 3,043 to 1,439, primarily due to the removal of documents with low cosine similarity. 
The number of unique words decreases from 29,931 to 3,857.
In the sentence version, we retain the same 1,439 documents, comprising 687,959 sentences after preprocessing. The further refined sentence subset consists of 678,218 sentences. The vocabulary sizes for these subsets are 5,789 and 5,766, respectively
\footnote{Literature provides limited information on the stepwise vocabulary reduction during preprocessing \citep{cai2022, bellstam2021, lowry2020, brown2020, garcia2023, bao2014, dyer2017, fernandes2020, mayasari2021, yau2014, farzadnia2024, albalawi2020, egger2022}. 
Previous studies report vocabulary sizes ranging from 2,000 to 8,000 words \citep{chen2017,chen2019, hong2018}, with \citet{jegadeesh2017} identifying over 60,000 unique words. 
However, their approach involved minimal preprocessing, omitting lemmatization or part-of-speech filtering. Consequently, we conclude that our vocabulary sizes fall within a reasonable range.}.

To systematically evaluate the impact of hyperparameter tuning, domain knowledge integration and input choice on our evaluation criteria, we employ a model for each input- (whole document, sentences, refined sentences), weighting- (tf, tfidf) and tuning-method (OTS, Tuned, Guided, Guided-Tuned\footnote{Due to lengthy computation times, Tuned LDA and HDP as well as Guided-Tuned LDA are applied exclusively to models based on whole-document inputs.}).
 
To establish a robust baseline, OTS models employ automatic detection of hyperparameters. An exception is the number of topics for LDA and NMF, which is set to 14, aligning with the number of topics from our keyword list. As HDP infers the number of topics from the data itself, no assumptions are necessary.
Moreover, we aim to evaluate and address the issues of topic instability and inefficient topic representation extraction \citep{agrawal2018, campbell2015}, by employing Tuned models.
This involves tuning all relevant hyperparameters to maximize Normalized Pointwise Mutual Information (NPMI) coherence (see \ref{sec:npmi_coherence}). We assume that the number of topics is at least 14, matching or exceeding the number of topics in our keyword list. 
For hyperparameter tuning of LDA, we opt for the Differential Evolution (DE) algorithm due to its capability of tuning multiple parameters simultaneously. Thus, we ensure an efficient and effective optimization \citep{fu2016}. 
To streamline the optimization process for NMF, we employ a grid search, as only the number of topics needs to be determined.
For HDP models, we follow a hybrid approach.
As HDP relies on integers to determine the first- and second-level concentration and truncation level, we initiate by performing a grid search over these four hyperparameters. After obtaining the coherence-maximizing combination, we apply a DE algorithm to identify the coherence-maximizing topic Dirichlet.
We evaluate the impact of incorporating domain knowledge by employing Guided models.
Specifically, we incorporate domain knowledge by using a $\eta$-matrix that assigns higher weight to topic-specific keywords. 
This matrix is designed to guide the models in recognizing relevant topics according to our predefined keyword list. 
Guided-Tuned models integrate both domain knowledge and hyperparameter tuning, aiming to align the models with domain-specific topics while simultaneously optimizing coherence for robust topic extraction.

For further improvement of the topic representations, we allow only Nouns to be included, as they are considered as the most representative topic words. Moreover, we ensure that none of the keywords are omitted from the topic representations.

\section{Results and Discussion - Classical Topic Modeling Approaches}
\label{sec:results_classical}
\label{sec:results_classical_tq}
\begin{table}[!ht]
\centering
\caption{Topic-Precision Scores - Classical Approaches.}
\label{Topic Precision Classical}
\begin{tabular}{lccc}
\toprule
 & \multicolumn{3}{c}{Type} \\ \cmidrule{2-4}
 Model & Whole document & Sentences & Refined Sentences \\ 
\midrule
\multicolumn{4}{l}{\textbf{OTS}} \\
LDA tf & 0 & 0.167 & 0.167 \\
LDA tfidf & 0.071 & 0.191 & 0.048 \\
HDP tf & 0 & 0.071 & 0 \\
HDP tfidf & 0 & 0.071 & 0 \\
NMF tf & 0 & 0.143 & 0.333 \\
NMF tfidf & 0.188 & 0.222 & 0.313 \\
\midrule
\multicolumn{4}{l}{\textbf{Guided}} \\
LDA tf & 0 & 0.071 & 0.214 \\
LDA tfidf & 0.511 & 0.262 & 0.405 \\
\midrule
\multicolumn{4}{l}{\textbf{Tuned}} \\
LDA tf & 0 & - & - \\
LDA tfidf & 0.071 & - & - \\
HDP tf & 0 & - & - \\
HDP tfidf & 0 & - & - \\
NMF tf & 0 & 0.441 & 0.339 \\
NMF tfidf & 0.188 & 0.375 & 0.289 \\
\midrule
\multicolumn{4}{l}{\textbf{Guided - Tuned}} \\
LDA tf & 0 & - & - \\
LDA tfidf & 0.786 & - & - \\
\bottomrule
\end{tabular}
\end{table}
Table \ref{Topic Precision Classical} presents the topic-precision scores of the classical topic modeling approaches.
Among OTS models, NMF performs best, particularly when using sentence-level input, achieving a topic-precision of 0.333.
LDA models follow, showing improved performance with sentence-based input.
HDP ranks last. Only with sentence input, HDP achieves a topic-precision of 0.071, while all other OTS configurations score zero.

Guiding the models enhances topic-precision in most cases, except for tf-weighted sentence-based input.
Overall, tf-idf weighting leads to better results, with whole-document input achieving the highest topic-precision of 0.511.

Tuning has a limited effect on whole-document input but improves results for sentence-based inputs.
However, even with these refinements, no tuned model surpasses the performance of guided LDA with tf-idf weighting on whole-document input.

Combining guidance and tuning yields the best topic-precision when using tf-idf weighting, while having no effect with tf weighting.
Guided and tuned LDA with tf-idf weighting achieves the highest topic-precision among classical approaches with a score of 0.786.
Yet, the model misses three financial topics, namely Liquidity, Financing and Tax and Regulation.
Hence, we conclude that none of the classical models reach the performance of contemporary topic modeling techniques, particularly when paired with FinTextSim (see Table \ref{Topic Precision Scores}).

\begin{table}[!ht]
\centering
\caption{Coherence Scores - Classical Approaches.}
\label{Coherence Scores Classical}
\begin{tabular}{lccc}
\toprule
 & \multicolumn{3}{c}{Type} \\ \cmidrule{2-4}
 Model & Whole document & Sentences & Refined Sentences \\ 
\midrule
\multicolumn{4}{l}{\textbf{OTS}} \\
LDA tf & 0 (0.008) & 0.009 (0.055) & 0.002 (0.011) \\
LDA tfidf & -0.172 & 0.004 (0.021) & 0 (0.002) \\
HDP tf & (-0.330) & (-0.327)  & (-0.330) \\
HDP tfidf & (-0.330) & (-0.327) & (-0.330) \\
NMF tf & 0 (0.031) & 0.025 (0.178) & 0.056 (0.168) \\
NMF tfidf & (-0.151) & 0.041 (0.185) & 0.062 (0.199) \\
\midrule
\multicolumn{4}{l}{\textbf{Guided}} \\
LDA tf & 0 (0.008) & 0.002 (0.029) & 0.006 (0.026) \\
LDA tfidf & (-0.045) & 0.006 (0.024) & 0.043 (0.017) \\
\midrule
\multicolumn{4}{l}{\textbf{Tuned}} \\
LDA tf & 0 (0.023) & - & - \\
LDA tfidf & 0.006 (0.083) & - & - \\
HDP tf & (-0.328) & - & - \\
HDP tfidf & (-0.328) & - & - \\
NMF tf & 0 (0.031) & 0.089 (0.201) & 0.069 (0.203) \\
NMF tfidf & (-0.046) & 0.093 (0.247) & 0.073 (0.251) \\
\midrule
\multicolumn{4}{l}{\textbf{Guided - Tuned}} \\
LDA tf & 0 (0.008) & - & - \\
LDA tfidf & 0.049 (0.063) & - & - \\
\bottomrule
\end{tabular}
\end{table}

Table \ref{Coherence Scores Classical} presents the coherence scores for classical topic modeling techniques, with non-topic-precision-weighted NPMI coherence shown in parentheses.
Incorporating topic-precision weighting reveals a critical weakness across all classical models: they fail to generate precise and meaningful topics within our financial dataset. 
Even the best-performing models exhibit severely diminished coherence when accounting for topic quality, demonstrating that none are truly effective for our domain.
While tuning and guiding approaches improve raw coherence scores, they fail to address the core issue of generating topics with economic relevance and precision. 
Even refined-sentence based tfidf-weighted NMF, which achieves the highest raw coherence, remains unreliable when topic-precision is considered. This highlights a fundamental limitation of classical techniques in handling financial disclosures.

Neglecting topic-precision weighting, NMF with sentence-based input achieves the highest coherence.
NMF's superior performance indicates its suitability for non-mainstream text, producing more coherent topics than LDA and HDP for our domain-specific dataset \citep{ocallaghan2015}.
LDA produces topics with lower coherence than NMF, consistent with \citet{egger2022,ocallaghan2015,chen2019}, but contrary to observations of \citet{albalawi2020} and \citet{farzadnia2024}.
HDP ranks last in terms of topic coherence, coinciding with \citet{mayasari2021} and \citet{farzadnia2024}. 
Contrary to \citet{altaweel2019}, our results do not support the assertion that both HDP and LDA are capable of finding good topics, particularly for HDP.
For whole-document OTS models, tf-weighting consistently yields higher coherence than tfidf-weighting. 
LDA achieves the highest coherence, followed by NMF, while HDP remains ineffective.
Using sentence-based input for OTS models, we find varying impacts: NMF and LDA show a notable enhancement, whereas HDP's coherence grows only slightly. The effect on NMF and LDA is even more significant for tfidf-weighting. For both weighting methods, NMF continues to outperform the other classical OTS models. The significant leap for NMF shows its superior performance for short-text modeling, aligning with \citet{chen2019}.
Further refining input sentences of OTS models leads to growing coherence only for tfidf-weighted NMF. All other models face a reduction. Based on the observed reduction of coherence scores between regular sentence-based and refined sentence-based models, we find that the models not only handle noise but leverage it to increase topic coherence. We assume that the increased sparsity of the input, induced by tfidf-weighting, helps the topic allocation, eventually improving coherence \citep{lee1999}.
Tuning the models leads to increased coherence scores with more significant improvement observed for tfidf-weighting.
We attribute this higher leverage to the nature of the financial vocabulary. In Finance, topic words tend to occur frequently within documents. Hence, tfidf-weighting downweighs their importance, resulting in increased sparseness and improved allocation of topics \citep{lee1999}. 
Guiding the models has only marginal effects on coherence, with the strongest impact observed for tfidf-weighted LDA. Hence, we only partially concur with \citet{chen2019} that incorporating domain knowledge enhances coherence.
Incorporating domain knowledge alongside hyperparameter tuning results in reduced coherence compared to purely tuned LDA models.
This is contrary to our initial expectations; Guided-Tuned LDA has lower coherence than tuned LDA. 
We assume that guiding the model restricts the range of possible hyperparameters, limiting optimization capabilities.

Overall, we conclude that BERTopic models significantly outperform classical models (see Tables \ref{Coherence Scores Classical} and \ref{Coherence Scores}), aligning with the findings of \citet{abuzayed2021} and \citet{egger2022}.
While non-weighted coherence scores suggest that tuned tfidf-weighted NMF performs best among classical models, incorporating topic-precision reveals their fundamental limitations. No classical approach is capable of generating both precise and coherent topics for financial disclosures.\footnote{Wordclouds for a selection of models are displayed in \ref{sec:classical_wordclouds}.} The reliance on modern techniques such as BERTopic is, therefore, essential for producing reliable financial topic models.

Given that classical models fail to produce meaningful economic topics consistently and reliably, we shift our focus toward enhancing contemporary topic modeling techniques. 
To this end, we introduce FinTextSim, designed to capture the unique characteristics of financial language and improve topic modeling for the financial domain.
Furthermore, as the topics generated by classical models are inherently weak, their separation becomes irrelevant. 
Consequently, we deliberately exclude the comparisons of topic similarity for classical models.

\section{Excerpt of Keyword List}
\label{sec:keyword_list}
\begin{itemize}
    \item Sales: sale, revenue, market, consumer, demand, competition, pricing, contract, price 
    \item Cost: cost, expense, liability, goodwill, impairment, depreciate, depreciation
    \item Profit/Loss: profit, performance, result, margin, income, earnings, loss, management, ebitda, ebda, ebit
    \item Operations: operations, production, business, produce, supply, operational, producer, process, processing, manufacturing, manufacture, supplychain, logistics, transport, marketing, advertising, advertisement, advertise 
    \item Liquidity: liquidity, interest, coverage, cash, capital, balance, cashflow, excess, working\_capital, work\_capital, excess\_cash
    \item Investment: investment, expenditure, m\&a, divestiture, invest, asset, disposal, divestment, 
    \item Financing: financing, finance, debt, equity, dividend, repurchase, share, funding, security, indebtness, indebtedness, borrowing, credit
    \item Litigation: litigation, lawsuit, legal, matter, dispute, complaint, arbitration, patent
    \item HR: employee, retention, hiring, hire, union, consultant, staff, recruiting, recruit, recruitment, labor, incentive, insurance, team, training, salary, wage, job, work
    \item Regulation: regulations, tax, government, tax\_expense, government\_affair, legislation, federal, regulator, regulate
    \item Accounting: accounting, account, audit, auditing, control, adjustment, filing, auditor, report
    \item Energy: energy, coal, solar, fuel, wind, water, electric, oil, megawatts, mwh, megawatthours, kilowatts, kwh, kilowatthours, gigawatts, gwh, gigawatthours
    \item ESG: plastic, recycle, waste, carbon, emission, renewable, environment, sustainable, sustain, ecologic, ecological
    \item Covid-19: covid, cov-19, pandemic, disease, corona, covid-19, sars-cov
\end{itemize}

\section{Topic Representations - Classical Approaches}
\label{sec:classical_wordclouds}

\begin{figure}[H]
\centering
\includegraphics[width=\textwidth]{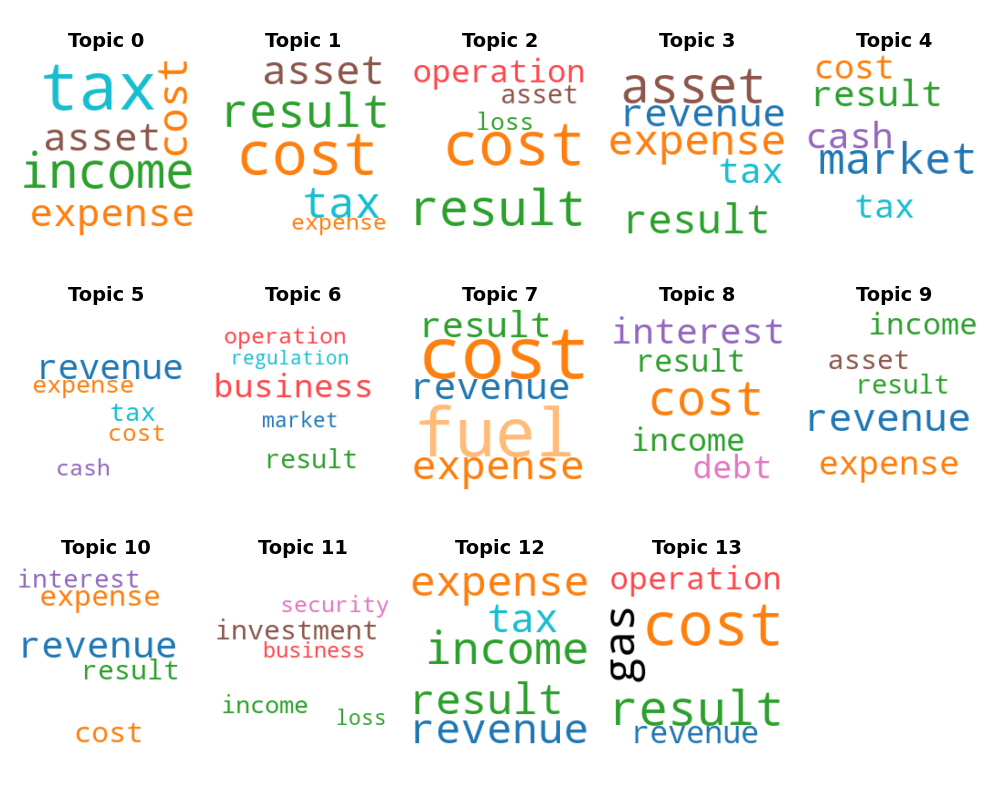}
\caption{Wordcloud - LDA - Whole Document - tf-weighting.\\ The color of each word represents its associated unique topic from the keyword list. Words colored in black are not present in the keyword list.}
\label{fig:wordcloud_lda_model}
\end{figure}

\begin{figure}[H]
\centering
\includegraphics[width=\textwidth]{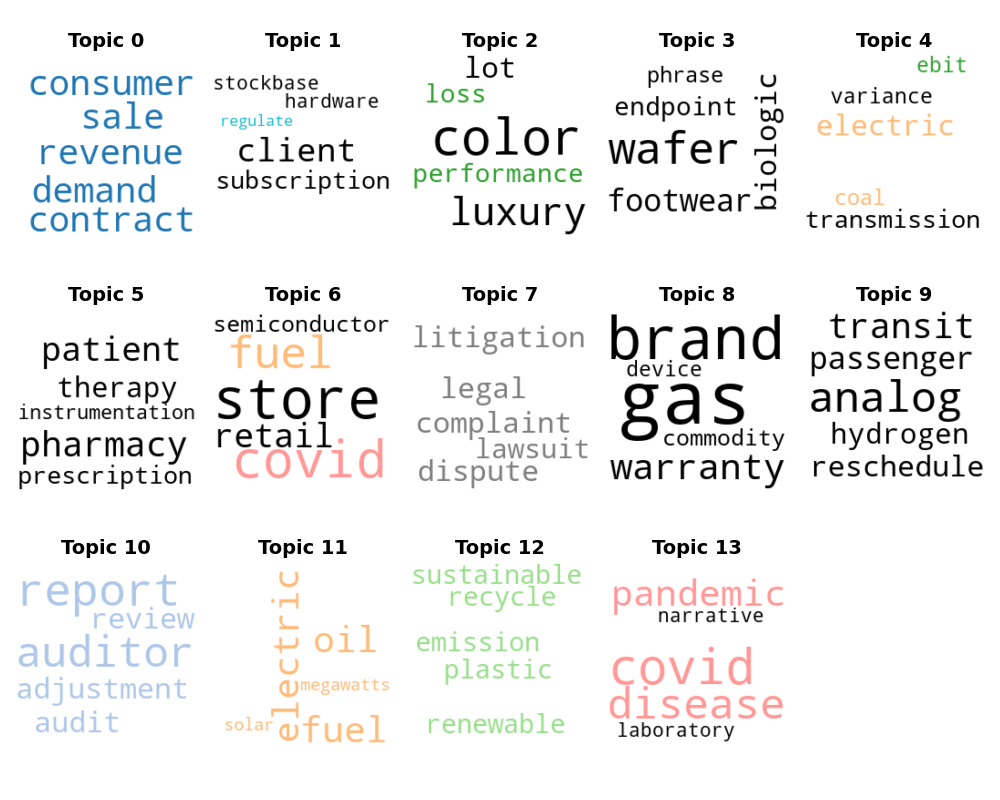}
\caption{Wordcloud - LDA - Whole Document - Guided - tfidf-weighting.\\ The color of each word represents its associated unique topic from the keyword list. Words colored in black are not present in the keyword list.}
\label{fig:wordcloud_lda_model_guided_tfidf}
\end{figure}

\begin{figure}[H]
\centering
\includegraphics[width=\textwidth]{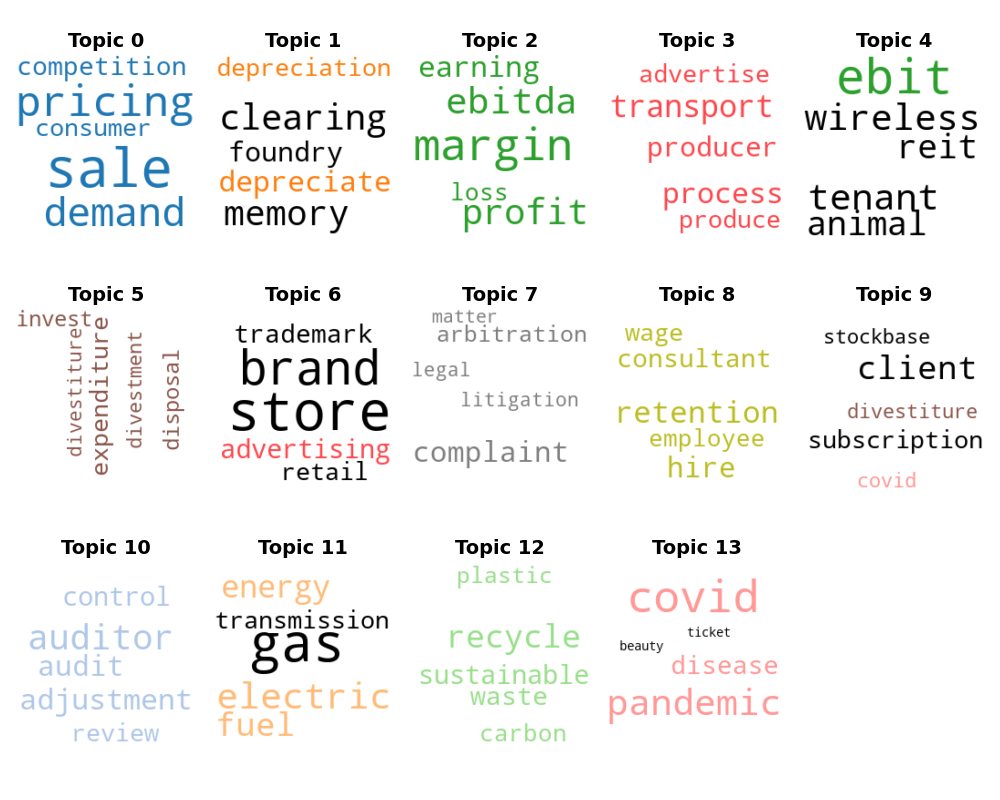}
\caption{Wordcloud - LDA - Whole Document - Guided-Tuned - tfidf-weighting.\\ The color of each word represents its associated unique topic from the keyword list. Words colored in black are not present in the keyword list.}
\label{fig:wordcloud_lda_guided_tuned_tfidf}
\end{figure}

\begin{figure}[H]
\centering
\includegraphics[width=\textwidth]{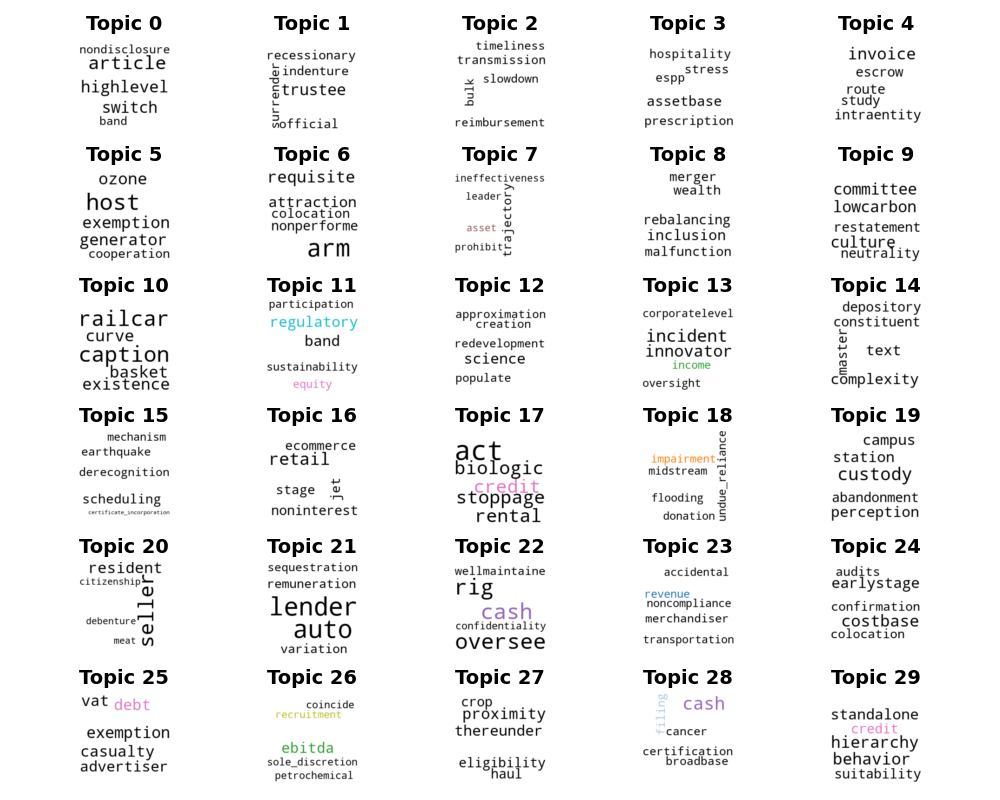}
\caption{Wordcloud - HDP - Whole Document - Tuned - tf-weighting.\\ The color of each word represents its associated unique topic from the keyword list. Words colored in black are not present in the keyword list.}
\label{fig:wordcloud_hdp_tuned_tfidf}
\end{figure}


\begin{figure}[H]
\centering
\includegraphics[width=\textwidth]{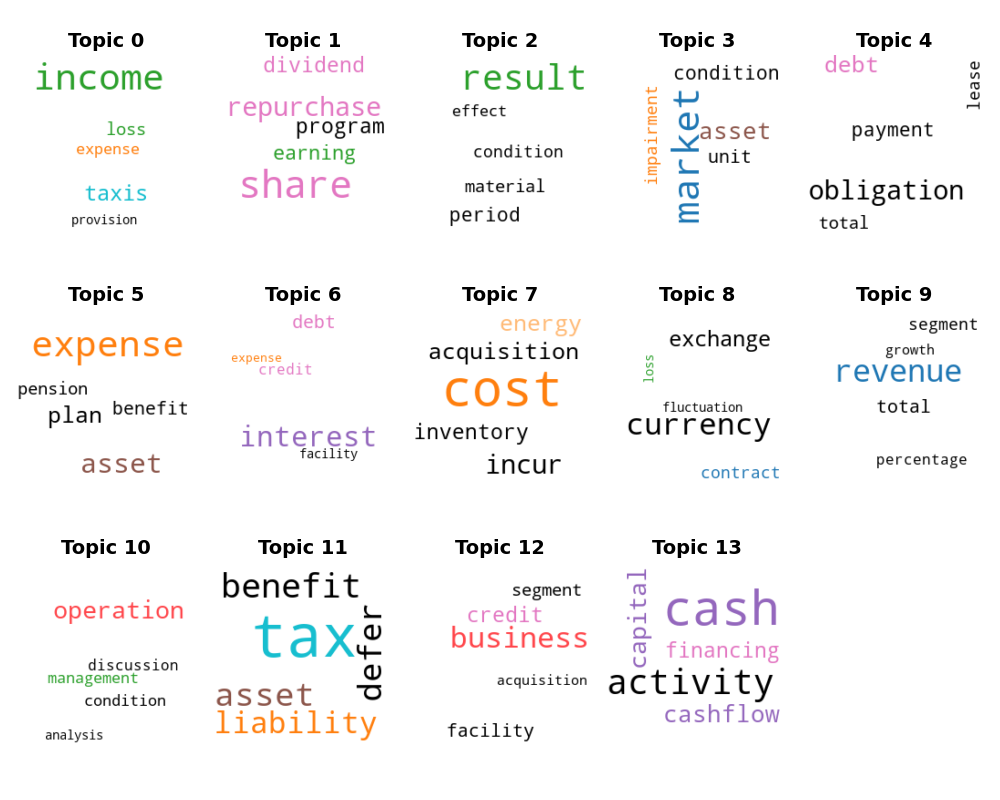}
\caption{Wordcloud - NMF - Sentences - tf-weighting.\\ The color of each word represents its associated unique topic from the keyword list. Words colored in black are not present in the keyword list.}
\label{fig:wordcloud_nmf_model_sentences}
\end{figure}

\begin{figure}[H]
\centering
\includegraphics[width=\textwidth]{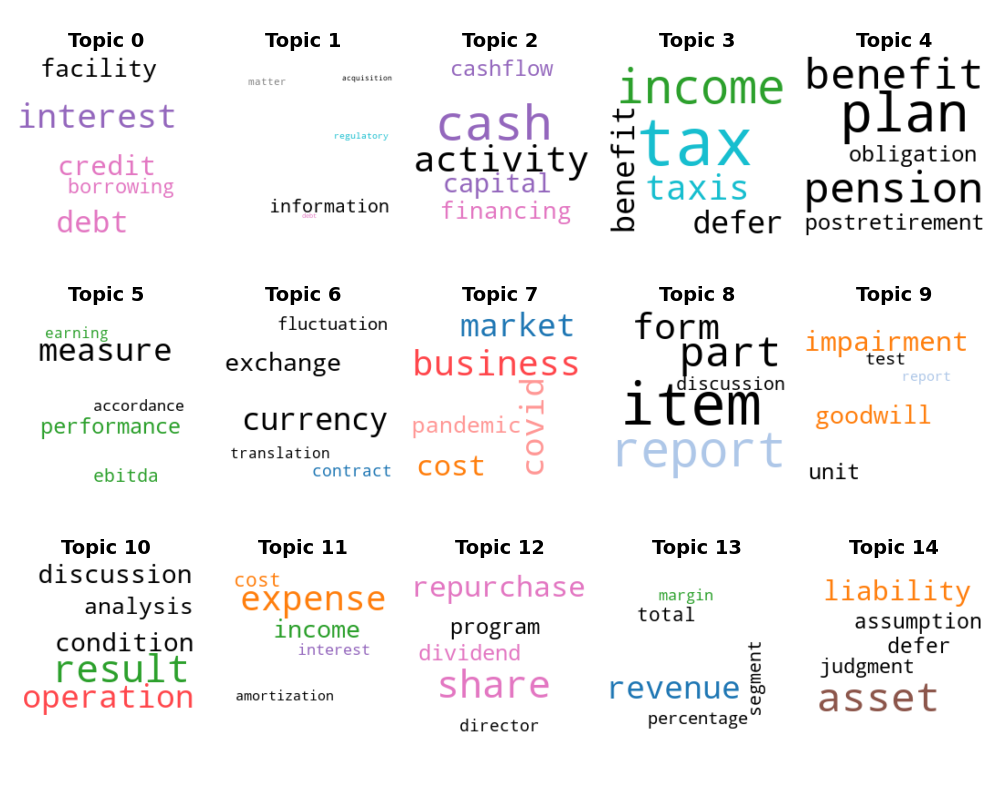}
\caption{Wordcloud - NMF - Sentences - Tuned - tfidf-weighting.\\ The color of each word represents its associated unique topic from the keyword list. Words colored in black are not present in the keyword list.}
\label{fig:wordcloud_nmf_sent_tuned_tfidf}
\end{figure}

\noindent 

\end{document}